\UseRawInputEncoding
\documentclass[a4paper]{article}
\usepackage{fullpage} 
\usepackage{parskip} 
\usepackage{tikz} 
\usepackage{amsmath}
\usepackage{hyperref}
\usepackage{svg}
\usepackage{tabularx}
\usepackage{booktabs}
\usepackage{gensymb}
\usepackage{Shorthands}
\usepackage{courier}
\usepackage{geometry}
\usepackage[labelfont=bf,format=plain,justification=justified,singlelinecheck=false]{caption}
\usepackage{lipsum}
\usepackage{url}
\usepackage{subfigure}
\usepackage{amsfonts}
\usepackage{cite}

\title{Real-time Dynamics of Soft Manipulators with Cross-section Inflation: Application to the Octopus Muscular Hydrostat
\thanks{
This work was supported by the Ministry of Education of Singapore, under the Tier 2 project REBOT - Rethinking Underwater Robot Manipulation (MOE-T2EP50221-0010), Khalifa University under Awards No.
RIG-2023-048, RC1-2018-KUCARS and in part by the US Office of Naval Research Global under Grant N62909- 21-1-2033. Github repo: \url{https://github.com/yuchensun97/sorosim_v7.0}
}}
\author{
    \parbox{\textwidth}{
    \small
      Yuchen Sun\textsuperscript{1}, Anup Teejo Mathew\textsuperscript{2,3}, Imran Afgan\textsuperscript{2}, Federico Renda\textsuperscript{2,3}, Cecilia Laschi\textsuperscript{1*} \\[1em]
      \small
      \textsuperscript{*} Corresponding Author, email: \texttt{mpeclc@nus.edu.sg} \\
      \textsuperscript{1}Department of Mechanical Engineering, National University of Singapore, Singapore, Singapore 117575 \\
      \small
      \textsuperscript{2}Department of Mechanical and Nuclear Engineering, Khalifa University, Abu Dhabi 127788, United Arab Emirates \\
      \small
      \textsuperscript{3}Khalifa University Center for Autonomous Robotics Systems (KUCARS), Khalifa University, Abu Dhabi 127788, United Arab Emirates
    }
}
\date{}

\renewcommand{\figurename}{Figure.}
\renewcommand{\tablename}{Table.} 
\newcommand{\eqname}{Eqn.} 
\newcommand{\sectioname}{Section.}

\newcommand{\brk}[1]{\left(#1\right)}
\newcommand{\code}[1]{\texttt{#1}}

\newcommand{\keywords}[1]{%
  \noindent\textbf{Keywords:} #1
}

\hypersetup{
    colorlinks=true,        
    linkcolor=blue,         
    filecolor=blue,         
    urlcolor=blue,          
    citecolor=blue,         
    pdftitle={arxiv},       
}

\begin{document}

\maketitle
\begin{abstract}
Inspired by the embodied intelligence of biological creatures like the octopus, the soft robotic arm utilizes its highly flexible structure to perform various tasks in the complex environment. While the classic Cosserat rod theory investigates the bending, twisting, shearing, and stretching of the soft arm, it fails to capture the in-plane deformation that occurs during certain tasks, particularly those involving active lateral traction. This paper introduces an extended Cosserat rod theory addressing these limitations by incorporating an extra strain variable reflecting the in-plane inflation ratio. To accurately describe the viscoelasticity effect of the soft body in dynamics, the proposed model enhances the constitutive law by integrating the Saint-Venant Kirchhoff hyperelastic and Kelvin-Voigt viscous models. The active and environmental loads are accounted for the equations of motion, which are numerically solved by adapting the Geometric Variable Strain (GVS) approach to balance the accuracy and computational efficiency. Our contributions include the derivation of the extended Cosserat rod theory in dynamic context, and the development of a reduced-order numerical method that enables rapid and precise solutions. We demonstrate applications of the model in stiffness tuning of a soft robotic arm and the study of complex octopus' arm motions. \\
\keywords{
Cosserat rod, Geometric Variable Strain, Soft Robotics, Embodied Intelligence}
\end{abstract}

\section{Introduction}
Soft robotics are largely inspired by the \textit{embodied intelligent} of biological creatures\cite{mengaldo_concise_2022, pfeifer2007self}: the physical body with compliant, elastic internal structure interacts continuously with the complex environment under the commands of nerve system. An example of embodied intelligence is the muscular hydrostats, a biological structure of muscle tissues with nearly infinite degrees of freedom, allowing the animal's body to perform a wide range of motions like bending, shortening, elongating, reaching, and stiffness tuning\cite{kier1985tongues, smith1989trunks}. A typical type of soft robot inspired by this principle takes the form of a slender beam, which includes designs such as the snake-like robot\cite{chirikjian_hyper-redundant_1994}, octupus-like manipulator\cite{calisti_octopus-bioinspired_2011, laschi_soft_2012}, and softworm\cite{10.5555/1036292}.  Compared with the rigid robots, which are of closed-form in finite joint space\cite{brady1982robot}, the inherent dexterity and continuity of soft robots makes the governing partial differential equations (PDEs) highly non-linear and subjected to infinite degrees of freedom (DoFs), and often precludes analytical solutions\cite{armanini_soft_2023}. Therefore, to describe the motion and deformation, and to exploit its embodied intelligence, a mathematical model that balances accuracy and numerical feasibility is required for further study of the robot\cite{10136424, lai2022constrained}. 

The most straightforward approach to model the dynamics of soft robots is through continuum 3D solid mechanics, which fully captures the complexity of soft materials\cite{mengaldo_concise_2022}. This model, which relies on the 3D FEM approach, can accurately simulates deformations in soft bodies, regardless of their initial shape or material properties. However, the large DoFs within the soft-body makes it expensive computationally and unsuitable for real-time simulation and control. The reduced-order modeling works\cite{webster_design_2010, jones2006kinematics, camarillo2008mechanics, camarillo_configuration_2009, chirikjian_closed-form_1995, zheng2012dynamic, stella2023piecewise} evolving from the rigid robot dynamics address this problem by a specific geometry approximation on the multi segments of the soft link. Such an assumption is valid only if the soft body is constrained to some particular strains \cite{armanini_soft_2023}, and thus, the geometrical model fails when there are complicated external loads applied to the soft link. 

The demands of no restrictions imposed on displacement or rotation lead to the adoption of \textit{Cosserat rod model}\cite{simo1985finite}. The classic Cosserat theory models a soft-slender beam as a one-dimensional structure consisting of an infinite sequence of infinitesimal rigid cross-sections, which enables the beam to show deformations with all six degrees of freedom--- stretching, shearing, bending, and twisting at the cross-sectional level. Previous study \cite{rucker_statics_2011} and \cite{renda_dynamic_2014}
apply the classic Cosserat theory to the soft robots subjected to complicated environment loads and thread-like actuation loads in statics and dynamics, respectively. In terms of the numerical solutions, the research \cite{gazzola_forward_2018, till_real-time_2019} predicts the time derivatives of the system's state variables based on the historic values of state variables and updates the rod's position and orientation along the length at each time step. On the other hand, the \textit{Geometric Variable Strain}(GVS) approach\cite{renda_discrete_2018, renda_geometric_2020, 9272318} projects the strain variables onto a finite joint space via a set of strain basis functions, and convert the original PDEs to a minimal set of ordinary differential equations (ODEs) in the form of Lagrangian mechanics. While previous works balance accuracy and computational cost, they do not fully capture the complexity of the soft body due to the rigid cross-sections assumption, where only deformations between consecutive infinitesimal cross-sections are considered, and thus the analysis of actuation loads are only limited to the axial or oblique actuators. If there are transversal actuators contracting the cross-section to stretch the soft rod-like \cite{mazzolai2012soft}, or the soft rod itself is a hollow tube, the classic Cosserat theory fails to remain geometrically exact. 
We also notice that transversal actuation is widely employed in a wide range of motions in creatures. For example, an octopus arm achieves elongation solely through the contraction of its transversal muscles and performs reaching motion by simultaneously releasing its oral longitudinal muscle while contracting the aboral and transversal ones\cite{kier_arrangement_2007}. 

Some studies have already been conducted to extend the Cosserat theory to address the limitations cited above. For example, in \cite{chang_energy_2023, shih2023hierarchical}, radial contraction is modeled as an external force along the rod's length. This method, while straightforward, does not provide a clear physical interpretation for this conversion and fails to reflect the cross-sectional deformation in computation. In contrast, research in \cite{kumar_geometrically_2011, tunay_spatial_2013} extend the classic Cosserat theory in \textit{static} contexts by introducing in-plane deformation strain variables and developing a new constitutive law derived from hyperelastic material models. In computer graphics, a study by Angles \cite{angles_viper_2019} applies the extended Cosserat theory to simulate the dynamics of musculoskeletal structures. For computational convenience, they simplify the constitutive law to a quadratic function of strain variables. While this simplification is practical for computer graphic simulation, it is important to note that it deviates from the physical laws that govern real-world soft materials. 

In this paper, we hope to create a model that is both efficient and robust enough for practical soft robots applications without sacrificing the detailed accuracy. In this work, for the first time, we derive an extended Cosserat theory in \textit{dynamics} that 
captures not only the deformation between infinitesimal cross-sections but also the in-plane inflation. This is done by introducing an extra strain variable reflecting the changes in magnitude of the body-fix director. To accurately describe the viscoelastic effect of the soft body in dynamics, we enhance the constitutive law by integrating the Saint-Venant Kirhhoff hyperelastic model and Kelvin-Voigt viscous model. We analyze all the external and internal loads applied to the soft link, including the environment loads, the longitudinal, oblique, and transversal actuation loads, and finally, obtain the equations of motion from momentum conservation law. We make the extended Cosserat theory computationally viable for soft robots application by parameterizing the extra strain variable in the GVS approach. The contributions of this work include:
\begin{itemize}
    \item Introduction of a new strain variable to capture the in-plane deformation of a soft rod, addressing the limitations of classic Cosserat theory.
    \item Derivation of an extended Cosserat rod theory in \textit{dynamics} context for viscoelastic materials from principles of three-dimensional continuum mechanics.
    \item Development of a reduced-order numerical method that enables both rapid and precise solutions for the dynamics of the soft rod.
    \item Implementation of our theory and numerical method in the study of creature motion and the actuation of soft robots.
\end{itemize}

This paper is organized as follows: \sectioname~\ref{sec: model} develops the mathematical model of the extended Cosserat rod with cross-section inflation from first principles. \sectioname~\ref{sec: load} discusses the possible external load applied on the soft body. \sectioname~\ref{sec: strain} proposes the strain parameterized method to reduce the order of the extended Cosserat rod theory for fast computation. \sectioname~\ref{sec:results} provides several applications using our method. Finally, \sectioname~\ref{sec:conclusion} concludes with a discussion of the limitations and future work.
\section{Mathematical Model}\label{sec: model}

\subsection{Kinematics of Deformation} \label{sec:kinematics}

The Cosserat theory of rods describes the dynamical behavior of a spatial slender rod undergoing large deformations in space by stretching, torsion, bending, and shearing\cite{antman_theory_2005}. In the classic Cosserat Rod theory, the cross-section of the rod is assumed to remain unchanged. Let $\{\be_1, \be_2, \be_3\}$ be the standard orthonormal basis for $\setE^3$. For a rod with total length of $L$, the configurations $\calB \subset \setE^3$ with material arclength $s \in [0, L] \subset \setR$ consists of the centroid of the rod $\br \in \setE^3$ and two orthonormal directors $\{\bd_1, \bd_2\} \subset \setE^3$ which spans the cross section $\Omega$ of the rod at the particular arclength $s$. Define the third director as $\bd_3 := \bd_1 \times \bd_2$ and thus $\{\bd_1, \bd_2, \bd_3\}$ forms another right-handed orthonormal basis for $\setE^3$. We call $\{\be_1, \be_2, \be_3\}$ as global frame and $\{\bd_1, \bd_2, \bd_3\}$ as local frame. The rotation matrix from the global frame to the local frame is defined as $\bd_i = \bR \be_i$, $\bR \in SO(3)$, $ i =\{1, 2, 3\}$.

Let the strain variables $\bnu \in \setR^3$ and $\bkappa \in \setR^3$ represent the linear strain and curvature in local frame along $s$, and thus they define the strain vector field of $\br$ and $\bR$: $\br' := \bR \bnu$, $\bR' := \bR \widetilde{\bkappa}$, 
where $\widetilde{(\cdot)}: \setR^3 \rightarrow \mathfrak{so}(3)$. 
The terms $\{\nu_1, \nu_2\}$ correspond to shear, and $\nu_3$ represents stretch, 
while $\{\kappa_1, \kappa_2\}$ correspond to bending, and $\kappa_3$ represents twisting.


Let $(X_1, X_2)$ denotes the coordinates of the subspace formed by director $\{\bd_1, \bd_2\}$. For any material point in $\calB$, we express its referenced position coordinates by 
$\Bar{\bp}(X_1, X_2, s): \setE^3 \rightarrow \setE^3$ as
$
    \Bar{\bp}(X_1, X_2, s) = \Bar{\br}(s) + X_\alpha\Bar{\bd_\alpha}(s)
$
where $\Bar{\br}$ and $\Bar{\bd_\alpha}$ represent the referenced centroid and cross-sectional directors, and the subscript $\alpha$ denotes summation running from 1 to 2, respectively. Similarly, the position vector of the deformed material point $\bp(X_1, X_2, s): \setE^3 \rightarrow \setE^3$ can be written as
$
    \bp(X_1, X_2, s) = \br(s) + X_\alpha\bd_\alpha(s)
$.

\begin{figure}[!tb]
    \centering
    \includegraphics[width=.8\textwidth]{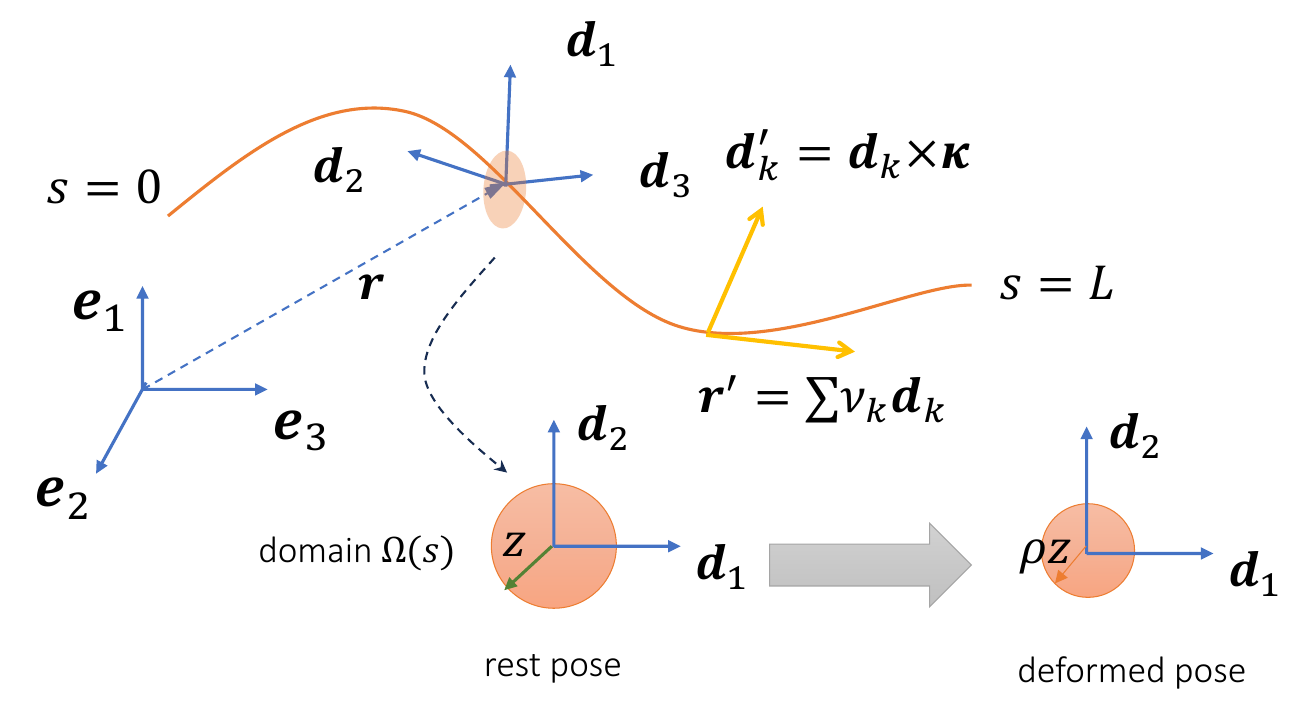}
    \caption{The kinematics of deformation of the extended Cosserat rod model. The geometric of the cross section is required to be axisymmetric. For a circular cross section, $z$ is the radius in referenced configuration}
    \label{fig: kinematics}
\end{figure}

To further reflect the exact shape of the slender rod, especially the homogeneous in-plane deformation, we 
introduce a new variable $\rho: s\in [0, L] \rightarrow (0, +\infty)$ representing the cross-sectional inflation ratio of axisymmetric geometry. In this case, the position of any referenced material point is rewritten as
$
    \Bar{\bp}(X_1, X_2, s) = \Bar{\br}(s) + \rho(s) X_\alpha\Bar{\bd_\alpha}(s)
$, and the position of the deformed material point is given by
$$
\bp(X_1, X_2, s) = \br(s) + \rho(s) X_\alpha\bd_\alpha(s) = \br(s) + \rho(s) X_\alpha\bR(s)\be_\alpha
$$
Figure \ref{fig: kinematics} illustrates the deformation of the extended Cosserat rod model. We assume no shear, stretch, bending, twisting, or inflation, i.e., stress-free in the referenced configuration in the following context so that 
$
\Bar{\bp}(X_1, X_2, s) = \begin{pmatrix}
    X_1 & X_2 & s
\end{pmatrix}^T
$.
Since the strain field variables are invariant to the in-plane inflation ratio $\rho$, the definition of $\{\bnu, \bkappa\}$ described in classic Cosserat rod theory still holds.

Now the deformation gradient tensor $\bfF \in \setE^{3 \times 3}$ is calculated by the gradient of $\bp$ (define as $\nabla \bp = \frac{\partial p_j}{\partial x_i}\be_j \otimes \be_i$) and its pseudo-polar decomposition\cite{geradin2001flexible}, resulting in
\begin{equation}
    \begin{aligned}
        \bfF = \nabla\bp &= \begin{pmatrix}
        \rho \bR \be_1 & | & \rho \bR \be_2  & |& \br' + X_\alpha(\rho \bR' + \rho' \bR)\be_\alpha 
    \end{pmatrix}\\
    &= \bR (\bD + \bI) = \bR\brk{\begin{pmatrix}
                        \rho-1  & 0 & a \\
                        0 & \rho-1 & b \\
                        0 & 0 & c-1
                        \end{pmatrix} + \bI}
    \end{aligned}
    \label{eqn:deformation tensor}
\end{equation}
which is measured by the local displacement tensor $\bD$ connected to the moving frame $\bR$, in terms of the strain variables $(\bnu, \bkappa)$ with components 
\begin{align*}
    &a(X_1, X_2) = \nu_1 + X_1 \rho' - X_2 \rho \kappa_3\\
    &b(X_1, X_2) = \nu_2 + X_1 \rho \kappa_3 + X_2 \rho' \\
    &c(X_1, X_2) = \nu_3 - X_1 \rho \kappa_2 + X_2 \rho \kappa_1
\end{align*}

The Green-Lagrangian strain tensor $\bE = \frac{1}{2}\brk{\bfF^T \bfF - \bI}$ is used to describe large deformation in continuum mechanics\cite{armanini_soft_2023}. Substituting \eqname~\ref{eqn:deformation tensor} into the formula results in the displacement expression of the strain tensor, we obtained
$
    \bE = \frac{1}{2}\brk{\bD^T + \bD + \bD^T\bD}
$
. Even though the rod is undergoing large deformation, the infinitesimal strain is still small compared to any other dimensions, such as material point displacement. Therefore, it is reasonable to linearize both strain and stress tensor at the undeformed configuration by discarding the quadratic term $\bD^T\bD$, providing that  $\norm{\bD}_F \ll 1$ is assumed to hold for a small strain, where $\norm{\cdot}_F$ denotes the Frobenius norm defined as $\norm{\bD}_F = \sqrt{\text{tr}(\bD^T \bD)}$, such that
\begin{equation}
    \bE = \frac{1}{2}\brk{\bD^T + \bD} = \begin{pmatrix}
        \rho-1 & 0 & \frac{1}{2}a \\
        0 & \rho-1 & \frac{1}{2}b \\
        \frac{1}{2}a & \frac{1}{2}b & c-1 
    \end{pmatrix}
    \label{eqn: small strain}
\end{equation}
\subsection{Constitutive Law with Kelvin-Voigt Viscous Damping}\label{sec: const law}

In 3D continuum mechanics, the elastic behavior can be specified in terms of a volume density of the strain energy function $W_E$\cite{smith_stable_2018}. Generally, the strain energy density function for the hyperelastic material might be well approximated by the quadratic function $W_E \approx \frac{1}{2}\bE:\setH:\bE$, where $\setH$ is the fourth order Hookean material tensor\cite{linn_geometrically_2013}. The corresponding stress tensor related to the strain and $W_E$ is called the second Piola-Kirchhoff (2nd PK) stress tensor, defined as $\bS_E = \partial W_E/\partial \bE \approx \setH : \bE$. In the case of homogeneous and isotropic material, $\setH$ depends on only two constant elastic moduli: the first and second Lam\'e parameters $\lambda$ and $\mu$, reducing the strain energy function to the Saint Venant-Kirchhoff (SVK) model as \eqname~\ref{eqn: wv} and \ref{eqn: 2nd pk}\cite{tunay_spatial_2013}.
\begin{align}
    &W_E = \frac{1}{2}\lambda \tr^2(\bE) + \mu \bE:\bE  \label{eqn: wv}\\
    &\bS_E = \lambda \tr(\bE)\bI + 2\mu \bE \label{eqn: 2nd pk}
\end{align}
If there is no shear, stretch, bending, twisting, and inflation in the referenced configuration, and the small strain assumption holds, then by substituting \eqname~\ref{eqn: small strain} into \eqname~\ref{eqn: wv} and \ref{eqn: 2nd pk}, we obtain the explicit form of the strain energy function and the 2nd Piola-Kirchhoff stress tensor:
\begin{gather*}
    W_E = \frac{\lambda}{2}\brk{2(\rho-1)+ (c-1)}^2 + \mu\brk{2(\rho-1)^2 + (c -1)^2+\frac{1}{2}a^2 + \frac{1}{2}b^2} \\
(\bS_E)_{ii} = 2(\lambda+\mu)(\rho-1)+ \lambda(c-1)\\
(\bS_E)_{ij} = 0, \quad i \neq j\\
(\bS_E)_{i3} = (\bS_E)_{3i} = 2\mu\bE_{3i}\\
(\bS_E)_{33} = 2\lambda(\rho-1)+(\lambda+2\mu)(c-1)
\end{gather*}
where $i, j\in\{1, 2\}$.



For isotropic incompressible material, Poisson's ratio $\nu^0 \rightarrow 0.5$, leading $\lambda \rightarrow +\infty$ and infinite large strain density energy. This issue can be addressed by noting that in isotropic materials, the normal in-plane strain $\bE_{ii}$ and the normal out-of-plane strain $\bE_{33}$ are related by $\bE_{ii} \equiv -\nu^0 \bE_{33}$. This indicates uniform in-plane inflation upon cross-section. Given that Poisson's ratio is related to $(\lambda, \mu)$ as $\nu^0 = \frac{\lambda}{2(\lambda+\mu)}$, we obtain $2(\lambda+\mu)(\rho-1) + \lambda(c-1) \equiv 0$, which implies that $(\bS_E)_{ii} = 0$, i.e. no lateral traction imposed. Integrating over the cross section, $2(\lambda + \mu)(\rho-1) + \lambda(\nu_3-1) \equiv 0$ always hold. If we let the constraint regarding the zero lateral traction holds, the strain energy density function is reduced to
\begin{equation}
    W_E = \frac{1}{2}\mu(a^2+b^2) +\frac{1}{2}E(c-1)^2
\end{equation}
where $E$ is the Young's modulus, defined as $E = \mu (3\lambda + 2\mu)/(\lambda+\mu)$.

The integration of $W_E$ over the cross-section domain $\Omega(s)$ yields the line density of strain energy function, that is
\begin{equation*}
    \Phi_E(s) = \int_{\Omega(s)} W_E(\bE) d\Omega
\end{equation*}
$\Phi_E(s)$ can be further relaxed by imposing $\rho\approx1$ in $a$, $b$, and $c$ so that only strain terms up to quadratic are retained. For a cross-section with axisymmetry, the simplest form of $\Phi_E$ is
\begin{equation}
    \begin{aligned}
        \Phi_E = 
        & \frac{1}{2} \mu A_0 (\nu_1^2 + \nu_2^2) + \frac{1}{2}E A_0 (\nu_3 - 1)^2 \\
        &+\frac{1}{2}EI_{\alpha\alpha}\kappa_\alpha^2 + \frac{1}{2}\mu I_{33} \kappa_3^2 + \frac{1}{2}\mu I_{33} \rho'^2
    \end{aligned}
    \label{eqn:Phi simple}
\end{equation}
where $A_0$ represents the referenced cross-section area and $I_{kk}$ represents the second moment inertia of $\Omega(s)$ about the director $\bd_k$. The Young's modulus $E$ can also be rewritten as
\begin{equation*}
    E = \lambda + 2\mu - \frac{\lambda^2}{\lambda + \mu}
\end{equation*}
Substituting $E$ into the second term of \eqname~\ref{eqn:Phi simple}, we obtain
\begin{equation*}
    \begin{aligned}
        \frac{1}{2}E A_0 (\nu_3 - 1)^2 &= \frac{1}{2}(\lambda + 2\mu)(\nu_3-1)^2 - \frac{1}{2}\frac{\lambda^2}{\lambda + \mu}(\nu_3 -1)^2 \\
        &= \frac{1}{2}(\lambda + 2\mu)(\nu_3-1)^2 + \frac{1}{2}\frac{\lambda^2}{\lambda + \mu}(\nu_3 -1)^2 - \frac{\lambda^2}{\lambda + \mu}(\nu_3 -1)^2 \\
        &= \frac{1}{2}(\lambda + 2\mu)(\nu_3-1)^2+ 2(\lambda + \mu)A_0 (\rho-1)^2 + 2\lambda A_0 (\rho-1)(\nu_3 - 1)
    \end{aligned}
\end{equation*}
In this way, we obtain a more complex expression of the in-plane strain energy function:
\begin{equation}
    \begin{aligned}
        \Phi_E = &2(\lambda + \mu)A_0 (\rho-1)^2 + 2\lambda A_0 (\rho-1)(\nu_3 - 1) \\
        & + \frac{1}{2} \mu A_0 (\nu_1^2 + \nu_2^2) + \frac{1}{2}(\lambda + 2\mu) A_0 (\nu_3 - 1)^2 \\
        &+\frac{1}{2}EI_{\alpha\alpha}\kappa_\alpha^2 + \frac{1}{2}\mu I_{33} \kappa_3^2 + \frac{1}{2}\mu I_{33} \rho'^2
    \end{aligned}
    \label{eqn:Phi}
\end{equation}
Although \eqname~\ref{eqn:Phi} may appear ad hoc, it can be justified in the analysis of internal force in the following context. In computation, the infinite large $\lambda$ for incompressible isotropic material is approximated by setting $\nu^0$ closed to 0.5.

When it comes to viscous damping, the simplest model is the Kelvin-Voigt damping, which is a linear combination of the stress and strain rate tensor. The dissipation rate per unit volume is given by $\Dot{D} = \Dot{\bE}: \setV: \Dot{\bE}$, where $\setV$ is the fourth order viscosity tensor. If the material is homogeneous and isotropic, the viscosity tensor totally depends on two constant parameters\cite{linn_geometrically_2013}: bulk viscosity $\zeta$ and shear viscosity $\eta$, such that $\Dot{D}$ is evaluated as
\begin{equation*}
    \Dot{D} = (\zeta - \frac{2}{3}\eta)\tr^2(\Dot{\bE}) + 2 \eta \Dot{\bE} : \Dot{\bE}
\end{equation*}
For incompressible material, constant volume indicates that $\tr^2(\Dot{\bE}) = 0$, so the first term vanishes. If we double integrate $\Dot{D}$ over the cross-section $\Omega(s)$ and time  and apply the same relaxing method, we obtain the quadratic line density of dissipation energy function in closed form for reflective symmetric cross-section:

\begin{equation}
    \begin{aligned}
        \Phi_D &= \int_0^t \int_{\Omega(s)} \Dot{D}d\Omega d\tau = \int_0^t \Dot{\Phi}_D d\tau\\
        & \approx \int_0^t \big( 4\eta A_0 \Dot{\rho}^2 +
        \eta A_0(\Dot{\nu}_1^2+\Dot{\nu}_2^2)\\
        &\quad \quad+ 2\eta A_0\Dot{\nu}_3^2  + 3\eta I_{\alpha\alpha}\Dot{\bkappa}_{\alpha}^2 \\
        & \quad \quad + \eta I_{33}\Dot{\bkappa}_3^2 + \eta I_{33}\Dot{\rho}'^2 \big) d\tau
    \end{aligned}
    \label{eqn:Phi D}
\end{equation}

The total energy per unit length is $\Phi = \Phi_E + \Phi_D$. To derive the relationship between $\Phi$ and the strain variables over $\Omega(s)$, we take the time derivative of $\Phi$ and apply the chain rule, yielding the total energy rate as:
\begin{equation}
    \begin{aligned}
        \Dot{\Phi} &= \int_\Omega \brk{\Dot{W}_E + \Dot{D}} d\Omega \\
        &= \frac{\partial \Phi}{\partial \bnu} \cdot \Dot{\bnu} +
        \frac{\partial \Phi}{\partial \bkappa} \cdot \Dot{\bkappa} +
        \frac{\partial \Phi}{\partial \rho} \Dot{\rho} +
        \frac{\partial \Phi}{\partial \rho'}\Dot{\rho}' \\
        &= \bn \cdot \bR \Dot{\bnu} + \bbm \cdot \bR \Dot{\bkappa} + Q \Dot{\rho}' + q\Dot{\rho}
    \end{aligned}
    \label{eqn: chain}
\end{equation}
where $\bn = \bR(\partial \Phi/\partial \bnu)$ and $\bbm = \bR(\partial \Phi/\partial \bkappa)$ are the stress and moment resultant per unit length, respectively, and $Q = \partial \Phi/\partial \rho'$ and $q = \partial \Phi/\partial \rho$ are the lateral traction resultants per unit length.

Alternatively, the total energy rate can also be given as the contraction of the first Piola-Kirchhoff stress tensor $\bP = \partial (W_E + D) / \partial \bF$ and the deformation rate tensor $\Dot{F}$, that is
\begin{equation}
    \begin{aligned}
        \Dot{\Phi} =& \int_\Omega \bP:\Dot{\bF} d\Omega = \bR \Dot{\bnu} \cdot \int_\Omega \bP \be_3 d\Omega +
        \bR \Dot{\bkappa} \cdot \int_\Omega X_\alpha \rho \be_\alpha \times \bP \be_3 d\Omega\\
        & +\be_\alpha \cdot \bR^T \Dot{\rho}'\int_\Omega X_\alpha \bP \be_3 d\Omega \\
        & + \be_\alpha \cdot \bR^T \Dot{\rho}\int_\Omega \bP \be_\alpha d\Omega - \be_\alpha \cdot \bkappa \times\bR^T \Dot{\rho} \int_\Omega X_\alpha \bP \be_3 d\Omega
    \end{aligned}
    \label{eqn: strain rate}
\end{equation}
The detailed steps to derive \eqname~\ref{eqn: strain rate} are listed in \ref{sec: strain energy rate}. By comparing \eqname~\ref{eqn: chain} and \eqname~\ref{eqn: strain rate}, we have
\begin{align}
    &\bn = \bR \frac{\partial \Phi}{\partial \bnu} = \int_\Omega \bP \be_3 d\Omega \label{eqn: n}\\
    &\bbm = \bR \frac{\partial \Phi}{\partial \bkappa} = \int_\Omega X_\alpha \rho \be_\alpha \times \bP \be_3 d\Omega \label{eqn: m}\\
    &Q = \frac{\partial \Phi}{\partial \rho'} = \be_\alpha \cdot \bR^T \int_\Omega X_\alpha \bP \be_3 d\Omega \label{eqn:Q}\\
    &q = \frac{\partial \Phi}{\partial \rho} = \be_\alpha \cdot \bR^T \int_\Omega \bP \be_\alpha d\Omega - \be_\alpha \cdot \bkappa \times\bR^T \int_\Omega X_\alpha \bP \be_3 d\Omega \label{eqn:q}
\end{align}

\begin{figure}[!tb]
    \centering
    \includegraphics[width=0.8\textwidth]{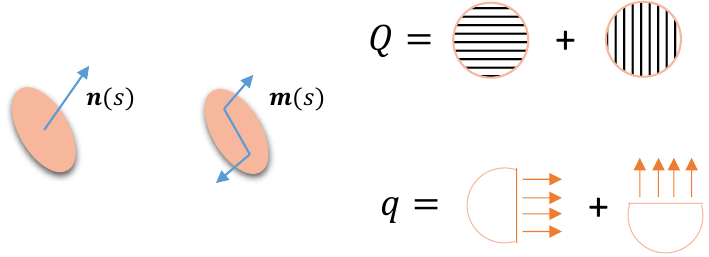}
    \caption{Physical interpretation of the stress, moment, and traction resultants. $\bn$ represents the stretch and shear force, $\bbm$ represents bending and twisting moments. $Q$ is the sum of $X_\alpha$-weighted integral of the shear traction components of $X_\alpha$, while $q$ is the sum of normal traction component along $X_\alpha$. All the resultants are measured on the cross-sectional level.}
    \label{fig: constitutive}
\end{figure}

 \eqname~\ref{eqn: n} indicates that $\bn$ is the stress resultant acting on $\Omega(s)$ in the reference configuration, while \eqname~\ref{eqn: m} indicates that $\bbm$ is the moment resultant bending $\Omega(s)$ along the rod axis. \eqname~\ref{eqn:Q} indicates that $Q$ is the $X_\alpha$-weighted shear traction resultant acting on $\Omega(s)$. Finally, \eqname~\ref{eqn:q} indicates that $q$ is the $X_\alpha$-weighted normal traction between two consecutive infinitesimal cross-sections. The physical interpretation of the above equations is straightforward, as shown in \figurename~\ref{fig: constitutive}.

Instead of the physical expressions, the explicit expressions of $\bn$, $\bbm$, $Q$, and $q$ derived from the strain derivative are used in real-time computation. We compute the strain derivatives from \eqname~\ref{eqn:Phi}, \ref{eqn:Phi D}, and let $\eta_E = 3\eta$, we have:
\begin{align}
    &\bn = \bR \brk{\bK_l(\bnu - \bnu^*)  + \bB_l \Dot{\bnu}+\bn^d}\label{eqn: n explicit}\\
    &\bbm = \bR \brk{\bK_\theta(\bkappa - \bkappa^*) + \bB_\theta \Dot{\bkappa}} \label{eqn: m explicit}\\
    &Q = \mu I _{33} \rho' + \eta I_{33} \Dot{\rho}'\label{eqn:Q explicit}\\
    &q = 4(\lambda + \mu) A_0 (\rho - 1) + 2 \lambda A_0(\nu_3-1) + 4\eta A_0\Dot{\rho}\label{eqn:q explicit}
\end{align}
where $\bK_l$, $\bK_\theta$ are the linear and angular stiffness matrices, $\bB_l$ and $\bB_\theta$ are the linear and angular damping matrices, $\bnu^*$ and $\bkappa^*$ are the reference configuration, $\bn^d$ is the elastic and damping stress resultant caused by in-plane deformation, respectively. The stiffness and damping matrices are listed as follows:
\begin{align*}
    \bK_l &= \diag\brk{\mu A_0, \mu A_0, (\lambda + 2\mu)A_0}\\
    \bK_\theta &= \diag\brk{E I_{11}, E I_{22}, \mu I_{33}}\\
    \bB_l &= \diag\brk{\eta A_0, \eta A_0, 2\eta A_0}\\
    \bB_\theta &= \diag\brk{\eta_E I_{11}, \eta_E I_{22}, \eta I_{33}}\\
    \bn^d &= \begin{pmatrix}
        0 & 0 & 2\lambda A_0 (\rho-1) 
    \end{pmatrix}^T
\end{align*}

The constitutive laws of classical Cosserat rod theory are derived using the same procedure, except that lateral tractions are not considered. Consequently, the constitutive laws are reduced to:
\begin{align}
    &\bn = \bR \brk{\bK_l(\bnu - \bnu^*)  + \bB_l \Dot{\bnu}}\label{eqn:classic n explicit}\\
    &\bbm = \bR \brk{\bK_\theta(\bkappa - \bkappa^*) + \bB_\theta \Dot{\bkappa}} \label{eqn:classic m explicit}
\end{align}
with
\begin{align*}
    \bK_l &= \diag\brk{\mu A_0, \mu A_0, E A_0}\\
    \bK_\theta &= \diag\brk{E I_{11}, E I_{22}, \mu I_{33}}\\
    \bB_l &= \diag\brk{\eta A_0, \eta A_0, \eta_E A_0}\\
    \bB_\theta &= \diag\brk{\eta_E I_{11}, \eta_E I_{22}, \eta I_{33}}
\end{align*}
\subsection{Dynamics}\label{sec:dynamics}

\begin{figure}[tb!]
    \centering
    \includegraphics[width=\linewidth]{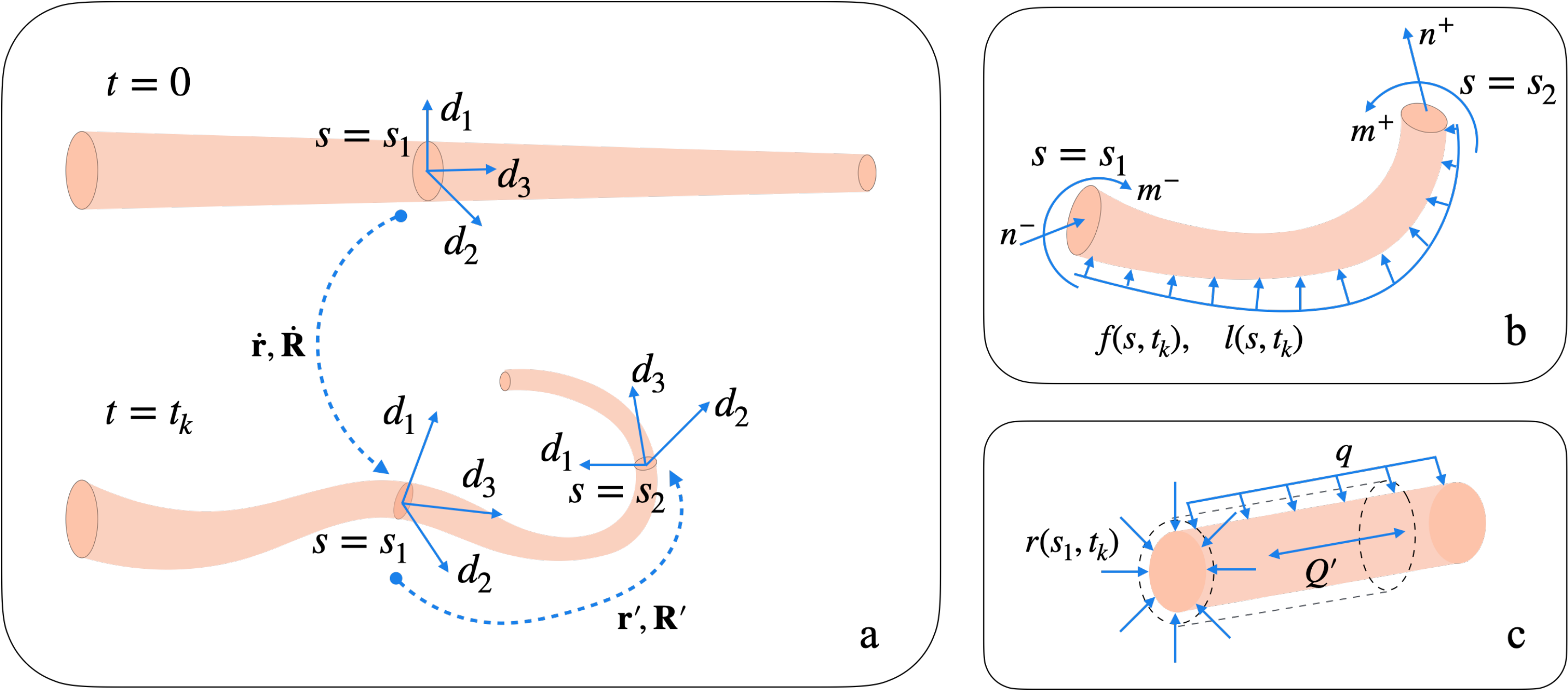}
    \caption{Schematics of momentum balance. a)compatibility law. Both the centroid's position and cross-sectional orientation are differentiable in time and space. b) linear and angular momentum balance of an arbitrary piece of the soft-slender. The internal forces $\bn$ and moments $\bbm$ are computed from \eqname~\ref{eqn: n explicit} and \ref{eqn: m explicit}, respectively. c) Momentum balance on consecutive infinitesimal cross sections, which is derived from material momentum balance. The dashed lines represent the referenced configuration.}
    \label{fig:momentum balance}
\end{figure}

To solve the 7 unknown strain variables of a soft beam with one end fixed to the wall, we require 7 equations derived from the momentum balance: three of them come from the linear momentum balance, another three are from angular momentum balance, and the last one comes from the material momentum balance. \figurename~\ref{fig:momentum balance} shows the schematics of the three momentum balance equations, together with the compatibility law, which is introduced later in \eqname~\ref{eqn:compact}.

Similar to the definition of strain variables, the linear velocity $\bu\in \setR^3$ and angular velocity $\bomega\in \setR^3$ in local frame at time $t$ define the velocity field of $\br$ and $\bR$: $\Dot{\br} = \bR \bu$, $\Dot{\bR} = \bR \widetilde{\bomega}$. Assuming $\br$ and $\bR$ are both continuously differentiable, then we have:
\begin{equation*}
    \frac{\partial^2 \br}{\partial t \partial s} = \frac{\partial^2 \br}{\partial t \partial s}, \quad \frac{\partial^2 \bR}{\partial t \partial s} = \frac{\partial^2 \bR}{\partial t \partial s}
\end{equation*}
Then the relation between strains and velocities is constructed as
\begin{equation}
    \bu' = \Dot{\bnu} + \bomega \times \bnu - \bkappa \times \bu,\quad
    \bomega' = \Dot{\bkappa} - \bkappa \times \bomega
    \label{eqn:compact}
\end{equation}

 Generally, the momentum balance can be stated as impulses of forces or moments equal to the linear or angular momentum change. Assume that at arclength $s$ and time $t$, the cross-sectional area is $A(s,t)$, and the matrix of the second moment of inertia is $\bI(s,t)$. Let $\bbf \in \setR^3$ and $\bl \in \setR^3$ be the external force and moment per unit length, and $\rho_0$ be the material density, then the linear momentum balance of the rod of segment $[s_1, s_2]$ from time $[0,t]$ is written as
\begin{equation}
    \begin{aligned}
        \int_0^t \brk{\bn^+(s_2,\tau) - \bn^-(s_1,\tau)} d\tau &+ \int_0^t \int_{s_1}^{s_2} \bbf(\xi, \tau) d\xi d\tau \\ 
    &= \int_{s_1}^{s_2}\rho_0\brk{A(\xi, \tau)\Dot{\br}(\xi,\tau) - A(\xi, 0)\Dot{\br}(\xi, 0)}d\xi
    \end{aligned}
    \label{eqn: dyn lmb}
\end{equation}

Similarly, we apply the angular impulse-momentum law and obtain
\begin{equation}
    \begin{aligned}
        &\int_0^t \big(\bbm^+(s_2,\tau) - \bbm^-(s_1,\tau) + \br(s_2,\tau) \times \bn^+(s_2,\tau) - \br(s_1,\tau)\times \bn^-(s_1,\tau)\big) d\tau \\
        &+ \int_0^t \int_{s_1}^{s_2} \big(\br(\xi, \tau) \times \bbf(\xi,\tau) + \bl(\xi,\tau)\big) d\xi d\tau \\
        &= \int_{s_1}^{s_2} \big(\rho_0 \brk{\bR(\xi,\tau) \bI(\xi,t) \bomega(\xi,\tau) - \bR(\xi, 0)\bI(\xi,0)\bomega(\xi,0)}\big)d\xi \\
        &+ \int_{s_1}^{s_2} \big(\br(\xi, \tau) \times \rho_0\brk{A(\xi, t)\Dot{\br}(\xi,t)}\big)d\xi
    \end{aligned}
    \label{eqn: dyn amb}
\end{equation}

By differentiating both sides of \eqname~\ref{eqn: dyn lmb}, \ref{eqn: dyn amb} w.r.t.$s$ and $t$, we obtain the linear and angular momentum balance equations as
\begin{align}
    &\bn' + \bbf = \rho_0 \bR \brk{\Dot{A} \bu + A \Dot{\bu} + \bomega \times (A\bu)} \label{eqn: dyn n}\\
    &\bbm' + \br' \times \bn + \bl = \rho_0 \bR \brk{\bomega \times \bI \bomega + \Dot{\bI}\bomega + \bI \Dot{\bomega}} \label{eqn: dyn m}
\end{align}
\eqname~\ref{eqn: dyn n} and \ref{eqn: dyn m} are of second order in the strain filed variables, requiring a set of 12 boundary conditions. In case of one end fixed to the wall and the other is subject to an applied force $\bF_l(t)$ and moment $\bL_l(t)$ at $s=L$, the rod meets the mixed boundary conditions, which can be stated as $\forall t\in(0, \tau]$, $\bnu(0, t) = \bnu^*(0)$, $\bkappa(0, t) = \bkappa^*(0)$,  $\bn(L, t) = \bF_l(t)$ and $\bbm(L,t)=\bL_l(t)$.

The third formula is given by the material momentum balance:
\begin{equation}
    \nabla \cdot \bP + \rho_0 \bb = \rho_0 \Ddot{\bp}
    \label{eqn: dyn 3db}
\end{equation}
where $\bb$ represents the body force per unit volume, and $\nabla \cdot (\cdot)$ operator denotes the divergence, which is defined as
$$
\nabla \cdot \bP = \frac{\partial \bP \be_\alpha}{\partial X_\alpha} + \frac{\partial \bP \be_3}{\partial s}
$$
By taking the derivative w.r.t $s$ of \eqname~\ref{eqn:Q}, we have
\begin{equation}
    \begin{aligned}
        Q' &= \be_\alpha \cdot \brk{-\bkappa \times \bR^T \int_\Omega X_\alpha \bP \be_3 d\Omega + \bR^T \int_{\Omega} X_\alpha \frac{\partial \bP \be_3}{\partial s}d\Omega} \\
        &= q - \be_\alpha \cdot \bR^T \brk{\oint_C X_\alpha \bP \cdot \bv dC + \int_\Omega X_\alpha \rho_0 \bb d\Omega}+ \be_\alpha \cdot \bR^T \int_\Omega X_\alpha \rho_0 \Ddot{\bp} d\Omega\\
        &= q - r + \be_\alpha \cdot \bR^T \int_\Omega X_\alpha \rho_0 \Ddot{\bp} d\Omega
    \end{aligned}
    \label{eqn: Q prime}
\end{equation}
where $r = \be_\alpha \cdot \bR^T \brk{\oint_C X_\alpha \bP \cdot \bv dC + \int_\Omega X_\alpha \rho_0 \bb d\Omega}$ captures the overall lateral traction acting on the cross-section $\Omega(s)$. $\bv$ is the outward normal vector to the reference boundary $C$ of $\Omega$. The first term of $r$ represents the force along the boundary of $\Omega(s)$ projecting to basis $\bR\be_\alpha$, while the second term represents the body force projected to the directors of $\Omega(s)$. If $\bb$ is homogeneous over $\Omega(s)$ (e.g. the only body force is gravity), the second term vanishes. The detailed steps to derive \eqname~\ref{eqn: Q prime} are listed in \ref{sec: traction}. By rearranging and expanding the last term of \eqname~\ref{eqn: Q prime}, we derive the last equation of motion:
\begin{equation}
    Q' - q + r = \rho_0 I_{\alpha\alpha}\brk{\Ddot{\rho} - \rho\widehat{\bomega}_{\alpha\alpha}^2}
    \label{eqn: dyn Q}
\end{equation},
where $\widehat{\bomega}_{\alpha\alpha}^2$ denotes the summation of the first two diagonal element of $\widehat{\bomega}^2$. \eqname~\ref{eqn: dyn Q} is of second order in the inflation ratio, requiring 2 boundary conditions in terms of $\rho$. There are three main types of boundary conditions commonly used, which are listed in \tablename~\ref{tab:BCs}.
\begin{table}[tb!]
    \centering
    \caption{Types of Boundary Conditions}
    \label{tab:BCs}
    \begin{tabular}{|c|c|c|} \hline 
         Type & Equations & Interpretation \\ \hline 
         Dirichlet & $\forall t\in (0, \tau]$, $\rho(0, t) = 1$, $\rho(L,t)=1$  & $\rho$ is fixed at both ends \\ \hline 
         Neumann & $\forall t\in (0, \tau]$, $Q(0, t) = 0$, $Q(L,t)=0$ & $\rho$ is free at both ends \\ \hline 
         Mixed & $\forall t\in (0, \tau]$, $\rho(0, t) = 1$, $Q(L,t)=0$ & $\rho$ is fixed at one end, and the other is free \\ \hline
    \end{tabular}
\end{table}
\eqname~\ref{eqn: dyn n}, \ref{eqn: dyn m} and \ref{eqn: dyn Q} with a total 14 sets of boundary conditions together form the dynamical equations of the Cosserat rod that allow cross-sectional inflation. All the stress resultants can be explicitly computed from the strain field variables (\eqname~\ref{eqn: n explicit}, \ref{eqn: m explicit}, \ref{eqn:Q explicit} and \ref{eqn:q explicit}), and the velocity field variables can be found by the compatibility law. For classic Cosserat rod theory, only the linear and angular momentum balance are considered, and the cross-sectional area and second moments of inertia are no longer variables of time and space.

\section{Internal Applied Forces and External Loads}\label{sec: load}
\begin{figure}[tb!]
    \centering
    \includegraphics[width=\linewidth]{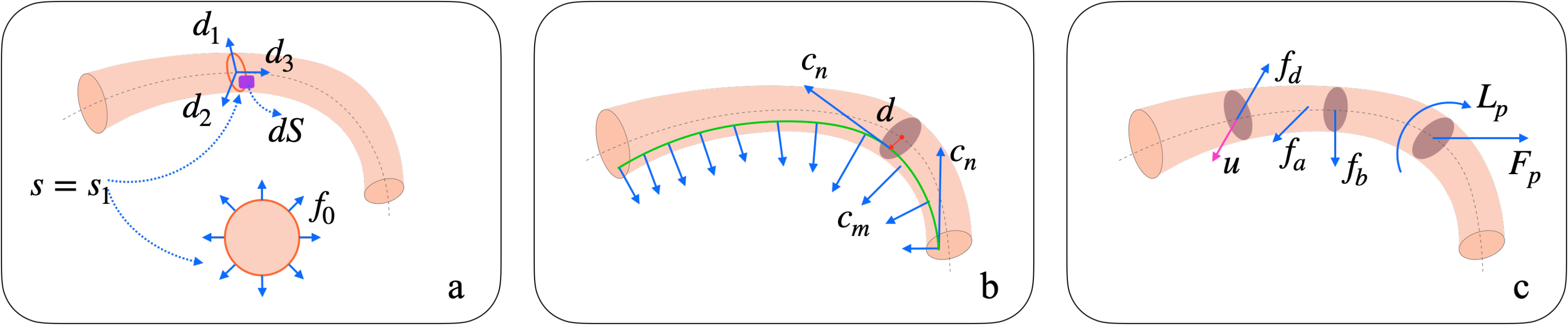}
    \caption{Internal applied force and external loads on the soft rod. (a) Transversal actuation loads are applied on the boundary of the cross-section and produce contraction or inflation traction. The orange line represents the transversal actuator. (b) Longitudinal or oblique actuation loads. This type of actuation produces forces tangential to the actuator route and moments proportional to the curvature of the actuator route. The green line represents the transversal or oblique actuator. (c) External loads including gravity and buoyancy $\bbf_b$, point force $\bF_p$ or moment $\bL_p$, dragging $\bbf_d$, and added-mass force $\bbf_a$ as the rod might interact with the environment}
    \label{fig:loads}
\end{figure}

The soft beam is driven by the internal applied force, such as the actuator's loads. In an underwater scenario, the soft beam is also affected by external forces like gravity, buoyancy, flow dragging, etc. The actuators are usually divided into two types\cite{laschi_soft_2012}: a) transversal actuator, which generates stress perpendicular to the longitudinal axis of the manipulator and on the boundary of the cross-sectional domain $\Omega(s)$. We assume that the magnitude of the stress is homogeneous over $\Omega(s)$; b) longitudinal or oblique actuator, which generates force with a decomposition along the length of the arm, and the resulting wrenches are distributed continuously at $s \in [x, y] \subset [0, L]$. \figurename~\ref{fig:loads} shows the important loads that might applied to the soft rod.

\subsection{Transversal Actuation Load}

We assume that the body force is homogeneous for all the material points in $\calB$, then we find that term $\int_\Omega X_\alpha \rho_0 \bb d\Omega$ of $r$ vanishes. Therefore, the lateral traction is only given by the force applied to the normal of the boundary $C$ of $\Omega(s)$. In the first term of $r$, $\bP \cdot \bv$ corresponds to the traction acting on the undeformed infinitesimal surface $dS$, which is normal to the boundary vector $\bv$, as shown in \figurename~\ref{fig:loads}(a). It is also a measurement of the internal force acting on $dS$. If there is no shear pressure acting on the boundary of $\Omega(s)$, and the transversal actuators exert homogeneous pressure $f_0^b$ which points inward to the origin of the deformed configuration in local frame, then we have $ \bP \cdot \bv dC = f_0^b  \bR \bv^b dC^b$, where $\bv^b$ is the outward normal vector to the deformed boundary $dC^b$. By multiplying the rotation matrix, we project the actual "force" from local to global frame (noticed that $\bR\bv^b \neq \bv$). Therefore, the lateral traction resultant is given by:
$$
r = \be_\alpha \cdot \bR^T \brk{\oint_C X_\alpha \bP \cdot \bv dC } = \be_\alpha \cdot \oint_{C} X_\alpha \rho f_0^b \bv^b dC = 2f_0^b \rho \pi z^2
$$
where $z$ is the radius of the cross-section in referenced configuration.

Alternatively, we can also derive the lateral traction resultant by computing the active Piola-Kirchhoff stress tensor \cite{ambrosi_active_2012} with pressure $f_0$ defined in the reference configuration. In the reference configuration, the active stress $f_0$ acts on the boundary $C$ with homogeneous magnitude, by \eqname~4.10 in \cite{ambrosi_active_2012}, we have the active Piola-Kirchhoff stress tensor as:
$$
\bP_a = \frac{f_0}{\sqrt{I_4}}\bfF \bv \otimes \bv
$$
where $I_4$ is an invariant defined as $I_4 = \bfF \bv \cdot \bfF \bv$, and the operator $\otimes$ denotes the tensor product. By substituting $\bP = \bP_a$ into the above equation, we have the lateral traction resultant as
\begin{equation}
    r = 2 f_0 \pi z^2
    \label{eqn: tm}
\end{equation}

\subsection{Longitudinal or Oblique Actuation Load}
A longitudinal or oblique actuator acts on the soft rod $\calB$ by an applied tension $T(t)\in \setR$ and a distributed moment along the actuator route inside $\calB$. Following \cite{renda_geometric_2020}, we express the longitudinal or oblique actuation route as $\br_c(s) = \br(s) + \rho(s) \bR(s) \bd$, $s\in[x,y]\subset[0,L]$, where $\bd = (Y_1, Y_2, 0)^T \in \setE^3$ denotes the coordinates of the actuator within the local frame. The resultant force will have the same magnitude as the tension, and point toward the tangent $\bt$ of the route $\br_c$, while the resultant distributed moment is proportional to $T(t)$ and points toward the direction of $\bd \times \bt$. To show the explicit expression of the longitudinal and oblique actuation load, we take the derivative of $\br_c$ w.r.t arclength as
\begin{align*}
    \br_c' &= \br'+ Y_\alpha\rho\bR'\be_\alpha + Y_\alpha'\rho\bR\be_\alpha+ Y_\alpha \rho' \bR \be_\alpha \\
    &= \bR \underbrace{\begin{pmatrix}
        \rho & 0 & a(Y_1, Y_2) \\
        0 & \rho & b(Y_1, Y_2) \\
        0 & 0 & c(Y_1, Y_2) \\
    \end{pmatrix}}_{\bfF_c}\underbrace{\begin{pmatrix}
        Y_1' \\
        Y_2' \\
        1
    \end{pmatrix}}_{\bd_0'}
\end{align*}
so that the unit tangent within the local frame can be written as $\bt = \bR^T\br_c'/\norm{\bR^T\br_c'}=\bfF_c \bd_0' / \norm{\bfF_c \bd_0'}$. With regards to the resultant force and moment, they are given by
\begin{equation}
    \bn_a = T(t)\bt, \quad \bbm_a = T(t) \bd \times \bt 
    \label{eqn: lm}
\end{equation}

\subsection{External Loads}
If the rod interacts with water, the soft rod deforms due to hydrodynamic forces, which in turn is influenced by the state of the soft rod. To describe these forces, we integrate the simplified hydrodynamic model proposed by \cite{9486942}. This model allows for the decoupling of the interactions between the soft rod and the water. The overall external forces and moments per unit length are expressed as follows:
\begin{align}
    &\bbf_e = \bbf_b + \bbf_a +\bbf_d + \bbf_p \\
    &\bl_e = \bl_p
\end{align}
where the subscripts $b$, $a$, $d$, and $p$ represent the contributions from gravity and buoyancy, added mass from the water, drag, and point forces or moments, respectively.

Assuming that the density of the water is $\rho_a$ and a homogeneous material density in the cross-section, then we have the gravity and buoyancy per unit length as
\begin{equation*}
    \bbf_b = (\rho_0-\rho_a)A\bG
\end{equation*}
where $A$ is the cross-sectional area, $\bG$ represents the gravity acceleration in inertial frame, which equals $(-9.81, 0, 0)^T$m/s$^2$. Since gravity and buoyancy are homogeneous on $\Omega(s)$, there are no extra moments produced w.r.t. the centroid.

The acceleration of the surrounding water produces the inertial forces interacting with the soft rod. Such added mass load is obtained as
\begin{equation}
    \bbf_a = -\bM_a \Dot{\bu} = -\pi z^2 \rho_a \text{diag}(B_1, B_2, 0) \Dot{\bu}
\end{equation}
on a circular cross section, where $B_1$, $B_2$ are constant added-mass coefficients in the $X_1$, $X_2$ direction, respectively. The added-mass loads act normal to the surface of the rod, and when there is geometrical symmetry on the cross-section, the added-mass moments will cancel out.

The dragging and lifting forces act opposite to the linear velocity of the cross-section and are proportional to its magnitude and the circular cross-sectional radius, given by
\begin{equation}
    \bbf_d = -\calD \norm{\bu} \bu = -z \rho_a \begin{pmatrix}
        C_D & -C_L & 0\\
        C_L & C_D & 0 \\
        0 & 0 & 0
    \end{pmatrix} \norm{\bu} \bu
\end{equation}
where $C_L$ and $C_D$ are lifting coefficient along the $\bd_3$ direction, and dragging coefficient acting on the $\bd_1$ and $\bd_2$ direction, respectively. The dragging load produces no moments for the geometrical symmetric cross-section.

When the rod is subjected to the point force $\bF_p$ and moment $\bL_p$ applied on $\Bar{s}$, we need to determine how the localized loads affect the rod.  We assume that both $\bF_p$ and $\bL_p$ concentrate on the origin of the cross-section $\Omega(s)$, then we can compute the distributed force $\bbf_p$ and moment $\bl_p$ per unit length as
\begin{equation}
    \bF_p(s) = \int_0^L \delta(s-\Bar{s})\bbf_p, \quad
    \bL_p(s) = \int_0^L \delta(s-\Bar{s})\bl_p
\end{equation}
respectively, where $\delta(s)$ is the dirac distribution. To be simplified, we take the value of $\bbf_p$ and $\bl_p$ the same as $\bF_p$ and $\bL_p$, except that the unit changes to N/m.

\section{Strain Parameterization} \label{sec: strain}
Fast computation requires reducing \eqname~\ref{eqn: dyn n}, \ref{eqn: dyn m}, and \ref{eqn: dyn Q} from continuous to a finite domain by parameterization. An effective approach is to extend the Geometric Variable Strain (GVS) model \cite{renda_geometric_2020} by parameterizing the extra strain variable $\rho$, such that \eqname~\ref{eqn: dyn n}, \ref{eqn: dyn m} and \ref{eqn: dyn Q} are projected to the general coordinates on the strain field.
The extended GVS model framework allows the efficient estimation of geometric Jacobian, statics, and dynamics of the system. The estimations are completed by a recursive computation of the forward kinematics, velocities, and acceleration.

\subsection{Recursive Kinematics}
The configuration of the soft rod $\calB$ w.r.t the inertial frame can be expressed using the space curve
\begin{equation}
    \bg(\cdot):s\in[0,L] \mapsto
    \bg(s) = \begin{pmatrix}
        \bR & \br \\
        0 & 1
    \end{pmatrix}\in
    \textit{SE}(3)
    \label{eqn: se3}
\end{equation}
and the inflation ratio of the cross-section $\rho$. By taking the derivatives of \eqname~\ref{eqn: se3} w.r.t $s$ and time, we obtain
\begin{equation}
    \bg'(s) = \bg(s) \widehat{\bxi}(s), \quad \Dot{\bg}(s)= \bg(s) \widehat{\bbbeta}(s)
\end{equation}
where
\begin{equation}
    \widehat{\bxi}= \begin{pmatrix}
        \widetilde{\bkappa} & \bnu \\
        0 & 0
    \end{pmatrix}\in \mathfrak{se}(3), \quad
    \bxi = (\bkappa^T, \bnu^T)^T \in \setR^6
    \label{eqn:twits}
\end{equation}
represents the strain twist in the body frame, while
\begin{equation}
    \widehat{\bbbeta} = \begin{pmatrix}
        \widetilde{\bomega} & \bu \\
        0 & 0
    \end{pmatrix}\in \mathfrak{se}(3), \quad
    \bbbeta = (\bomega^T, \bu^T )^T \in \setR^6
    \label{eqn:velocities}
\end{equation}
represents the screw velocity in the body frame. Given the strain twists $\bxi$ at $s$, we obtain the centroid's configuration as an exponential map\cite{renda_geometrically-exact_2022, hairer_geometric_2013}
\begin{equation}
    \bg(s) = \text{exp}\brk{\widehat{\bOmega}(s)}, \quad \text{with } \bg(0) = \bI_4
\end{equation}
where $\widehat{\bOmega}$ is the Magnus expansion of $\bxi$ at $s$. The fourth-order Zannah quadrature approximation, which balances the accuracy and computation cost, is provided in \ref{sec:recursive}. By integrating the compatibility law given by \eqname~\ref{eqn:compact} along the length of the rod, we obtain 
\begin{equation}
    \bbbeta = \text{Ad}^{-1}_{\bg}\int_0^s \text{Ad}_{\bg} \Dot{\bxi} ds
\end{equation}
Next, we parameterized $\bxi$ using a polynomial basis and the general coordinates $\bq_\xi \in \setR^{n_\xi}$ of dimension $n_\xi$ as:
\begin{equation}
    \bxi(s, \bq_\xi) = \bPhi_\xi (s)\bq_\xi + \bxi^*(s)
\end{equation}
The basis $\bPhi_\xi \in \setR^{6\times n_\xi}$ is independent of $\bq_\xi$ and time $t$, and subjected to the constrained strains such as the boundary conditions. $\bxi^* \in \setR^6$ is the reference strain twist which models the unstressed configuration. 

For the variable strain case, the Magnus expansion $\widehat{\bOmega}(h)$ is approximated through small intervals $h$ of $s$, and the kinematic terms are computed recursively along the length of the rod.
\begin{align}
    \bg(s+h) &= \bg(s)\text{exp}\brk{\widehat{\bOmega}(h)} \\
    \bbbeta(s+h) &= \text{Ad}^{-1}_{\text{exp}\brk{\widehat{\bOmega}}}\brk{\bbbeta(s)+T_{\bOmega} \bPhi_{\bOmega} \Dot{\bq}_\xi} \label{eqn: recursive eta}
\end{align}
where $T_{\bOmega}$ is the tangent operator of the exponential map and $\bPhi_{\bOmega}$ is the Magnus expansion of the basis \cite{10.5555/1036292}. The geometric Jacobian $\bJ(\bq_\xi, s)\in \setR^{6\times n_\xi}$ maps "joint" velocities to the body velocities according to
\begin{equation}
    \bbbeta = \bJ \Dot{\bq}_\xi
    \label{eqn: vel}
\end{equation}
Substituting $\bJ$ into \eqname~\ref{eqn: recursive eta}, we obtain
\begin{equation}
    \bJ(\bq_\xi, s+h) = \text{Ad}^{-1}_{\text{exp}\brk{\widehat{\bOmega}}}\brk{\bJ(\bq_\xi, s)+T_{\bOmega} \bPhi_{\bOmega}}
\end{equation}
The acceleration twist is derived as
\begin{equation}
    \Dot{\bbbeta}(s, \bq_\xi, \Dot{\bq}_\xi, \Ddot{\bq}_\xi) = \bJ(s, \bq_\xi)\Ddot{\bq}_\xi + \Dot{\bJ}(s, \bq_\xi, \Dot{\bq}_\xi)\Dot{\bq}_\xi
    \label{eqn: acc}
\end{equation}
The detailed steps to perform recursive computation are listed in \ref{sec:recursive}.

Similarly, the inflation ratio $\rho$ is parameterized as
\begin{equation}
    \rho (s, \bq_\rho) = \bPhi_\rho(s) \bq_\rho + \rho^*(s)
\end{equation}
where the basis $\bPhi_\rho \in \setR^{1\times n_\rho}$ is independent of the general coordinate $\bq_\rho \in \setR^{n_\rho}$ and is subjected to constraints of $\rho$. $\rho^*$ is the reference inflation ratio under the unstressed configuration. Since we have $\Dot{\rho} = \bPhi_\rho \Dot{\bq}_\rho$, we notice that the basis $\bPhi_\rho$ is itself the Jacobian matrix for $\bq_\rho$ and $\rho$. The acceleration of $\rho$ is computed as $\Ddot{\rho} = \bPhi_\rho \Ddot{\bq}_\rho$.

\subsection{General Dynamic Equations}\label{sec: general eom}
The equations of motion \eqname~\ref{eqn: dyn n} and \ref{eqn: dyn m} can be rewritten in SE(3) as \cite{boyer2017poincare}
\begin{equation}
    \Bar{\bm{\calM}}\Dot{\bbbeta} + \Dot{\Bar{\bm{\calM}}}\bbbeta + \text{ad}^*_{\bbbeta}\Bar{\bm{\calM}}\bbbeta = \brk{\bm{\calF}_i - \bm{\calF}_a}'+ \text{ad}^*_{\bxi}\brk{\bm{\calF}_i - \bm{\calF}_a} + \Bar{\bm{\calF}}_e
    \label{eqn: gen dyn}
\end{equation}
with the boundary condition
\begin{align*}
    \bg(0, t) = \bI_4,\quad \forall t\in (0, \tau] \\
    \bm{\calF}_i(L, t) = 0, \quad \forall t \in (0, \tau]
\end{align*}
where $\Bar{\bm{\calM}}(s, t) \in \setR^{6\times6}$ is the inertia matrix of the cross section. $\Bar{\bm{\calM}}= \rho_0 \text{diag}\brk{I_{11}, I_{22}, I_{33}, A, A, A}$ if the cross section is reflective symmetric. $\bm{\calF}_i, \bm{\calF}_a \in \setR^6$ are the elastic and longitudinal actuation load, respectively. $\Bar{\bm{\calF}}_e \in \setR^6$ is the distributed external load actuation. Together with \eqname~\ref{eqn: dyn Q}, and one of the boundary conditions listed in \tablename~\ref{tab:BCs}, we obtain the complete set of equations of motion. For a soft rod following the constitutive law described in \sectioname~\ref{sec: const law}, the internal distributed load is written as
\begin{align}
    &\begin{aligned}
        \bm{\calF}_i(s) &= \bm{\Sigma}(\bxi - \bxi^*) + \bm{\Upsilon}\Dot{\bxi} + \bm{\sigma}(\rho-\rho^*)  \\
        &= \bm{\Sigma}\bPhi_\xi \bq_\xi + \bm{\Upsilon}\bPhi_\xi\Dot{\bq}_\xi + \bm{\sigma}\bPhi_\rho \bq_\rho \in \setR^6
    \end{aligned} \\
    &\begin{aligned}
        Q(s) = \mu I_{33}{\bPhi_{\rho}}'\bq_{\rho} + \mu I_{33}\rho'^{*} + \eta I_{33}{\bPhi_{\rho}}'\Dot{\bq}_{\rho} \in \setR
    \end{aligned} \label{eqn: gen Q} \\
    &\begin{aligned}
        q(s) = 4(\lambda + \mu) A_0 \bPhi_\rho \bq_\rho + 4\eta A_0\bPhi_\rho \Dot{\bq}_\rho + \bm{\Sigma}_\rho \bPhi_\xi \bq_\xi \in \setR
    \end{aligned} \label{eqn: gen q}
\end{align}
where
\[
\bm{\Sigma} = \begin{pmatrix}
    \bK_\theta & 0 \\
    0 & \bK_l
\end{pmatrix}\in \setR^{6\times 6}, \quad \bm{\Upsilon} = \begin{pmatrix}
    \bB_\theta & 0 \\
    0 & \bB_l
\end{pmatrix}\in \setR^{6\times 6}
\]
are the screw elasticity and damping matrix along the length of the rod, 
\begin{align*}
    \bm{\sigma} =& (0, 0, 0, 0, 0, 2 \lambda A_0)^T \in \setR^{6\times1} 
\end{align*}
is the screw elasticity caused by cross-sectional deformation
\begin{align*}
    \bm{\Sigma}_\rho =& (0, 0, 0, 0, 0, 2\lambda A_0) \in \setR^{1\times6} 
\end{align*}
is the inflation elasticity caused by longitudinal deformation, respectively.

In terms of the actuation load$\bm{\calF}_a$ and $r$, we can rewrite them using the actuation bases. For thread-like longitudinal, the actuation basis is given by
\begin{equation}
    \bm{\calF}_a (s) = \sum_{k=1}^{n_{a\xi}}\begin{pmatrix}
        \rho \widetilde{\bd}_k \bt_k \\
        \bt_k
    \end{pmatrix}u_k = \bPhi_a \bfu
\end{equation}
where $\bPhi_a (q_\xi, q_\rho, s, t) \in \setR^{6\times n_{a\xi}}$ is the longitudinal actuation basis, with $n_{a\xi}$ being the number of longitudinal actuators on the rod. For homogeneous transversal actuators, the actuation basis is given by
\begin{equation}
    r = \sum_{k=1}^{n_{a\rho}}2\pi z_k^2 f_{0k} = \bPhi_r \bff_0
\end{equation}
where $\bPhi_r(\bq_\rho, s, t) \in \setR^{1\times n_{a\rho}}$ is the transversal actuation basis, with $ n_{a\rho}$ being the number of longitudinal actuators on the rod.

Substituting \eqname~\ref{eqn: vel}, \ref{eqn: acc} into \eqname~\ref{eqn: gen dyn}, \eqname~\ref{eqn: gen Q}, \ref{eqn: gen q} into \eqname~\ref{eqn: dyn Q}, and projecting them onto $\bq_\xi$, $\bq_\rho$ space through the Jacobians $\bJ$ and $\bPhi_\rho$ respectively, then integrating along the length of the rod, we obtain the equations of motion in general coordinates space:
\begin{align}
    \bM_\xi \Ddot{\bq}_\xi + (\bC+\bD_\xi) \Dot{\bq}_\xi + \bK_\xi \bq_\xi + \Bar{\bK}_\xi \bq_\rho &= \bB_\xi\bfu + \bF_\xi 
     \label{eqn: 1st gen eom}\\
    \bM_\rho \Ddot{\bq}_\rho + \bD_\rho \Dot{\bq}_\rho + \bK_\rho \bq_\xi + \Bar{\bK}_\rho \bq_\xi &= \bB_\rho \bff_0 + \bF_\rho \label{eqn: 2nd gen eom}
\end{align} 
where $\bM_\xi \in \setR^{n_\xi \times n_\xi}$ and $\bM_\rho \in \setR^{n_\xi \times n_\xi}$ are the general mass matrices, $\bC \in \setR^{n_\rho \times n_\rho}$ is the Coriolis matrix, $\bD_\xi \in \setR^{n_\xi \times n_\xi}$ and $\bD_\rho \in \setR^{n_\rho \times n_\rho }$ are the general damping matrix, $\bK_\xi \in \setR^{n_\xi \times n_\xi}$ and $\bK_\rho \in \setR^{n_\rho \times n_\rho}$ are the general stiffness matrix, $\Bar{\bK}_\xi \in \setR^{n_\xi \times n_\rho}$ and $\Bar{\bK}_\rho \in \setR^{n_\rho \times n_\xi}$ are the general stiffness matrix caused by in-plane deformation, $\bB_\xi\in\setR^{n_\xi \times n_{a\xi}}$ and $\bB_\rho \in \setR^{n_\rho \times n_{a\rho}}$ are the general actuation matrix, $\bF_\xi \in \setR^{n_\xi}$ is the general external forces, $\bF_\rho \in \setR^{n_\rho}$ is the general external traction caused by angular velocity, and $\bfu\in \setR^{n_{a\xi}}$ and $\bff_0 \in \setR$. The explicit expressions of these parameter matrices are listed in \ref{sec: coeff}. For static problem, all time-dependent variables are eliminated, and therefore \eqname~\ref{eqn: 1st gen eom} and \ref{eqn: 2nd gen eom} are reduced to
\begin{align}
    \bK_\xi \bq_\xi + \Bar{\bK}_\xi \bq_\rho &= \bB_\xi\bfu + \bF_\xi 
     \label{eqn: 1st statics}\\
    \bK_\rho \bq_\rho + \Bar{\bK}_\rho \bq_\xi &= \bB_\rho \bff_0  \label{eqn: 2nd statics}
\end{align}
In \eqname~\ref{eqn: 1st gen eom}, $\bM_\xi$, $\bC$, $\bB_\xi$, and $\bF_\xi$ vary with time and space while other parameters remain constant. In \eqname~\ref{eqn: 2nd gen eom}, $\bK_\rho$ and $\bF_\rho$ are time-dependent, with $\bK_\rho$ becoming constant in static cases; all other parameters are constant in both \eqname~\ref{eqn: 2nd gen eom} and \ref{eqn: 2nd statics}. The detailed steps for computation are listed in \ref{sec:computation}.

\section{Applications}\label{sec:results}
In this section, we provide three simulation examples of cable-driven bio-inspired soft manipulators to show the capabilities of our model. The soft manipulator can deform to complex shapes by applying both longitudinal and transversal tension to the actuator inside the soft body. In the first example, we focus on how the soft body tunes its stiffness by modulating longitudinal and transversal loads to resist both compression, unlike the classic model, which only allows for tension. In the remaining examples, we investigate the planar reaching and the 3D fetching motion of an octopus-like soft robotic arm, respectively. By comparing our model with the classic Cosserat model, we find that our model is more consistent with the biological profiles of a real octopus. This enhanced performance highlights the potential of our model in developing more efficient soft robotic systems.

\subsection{Axial Stiffness Tuning}
\begin{figure}[tb!]
    \centering
    \subfigure[]{
    \includegraphics[width=.7\linewidth]{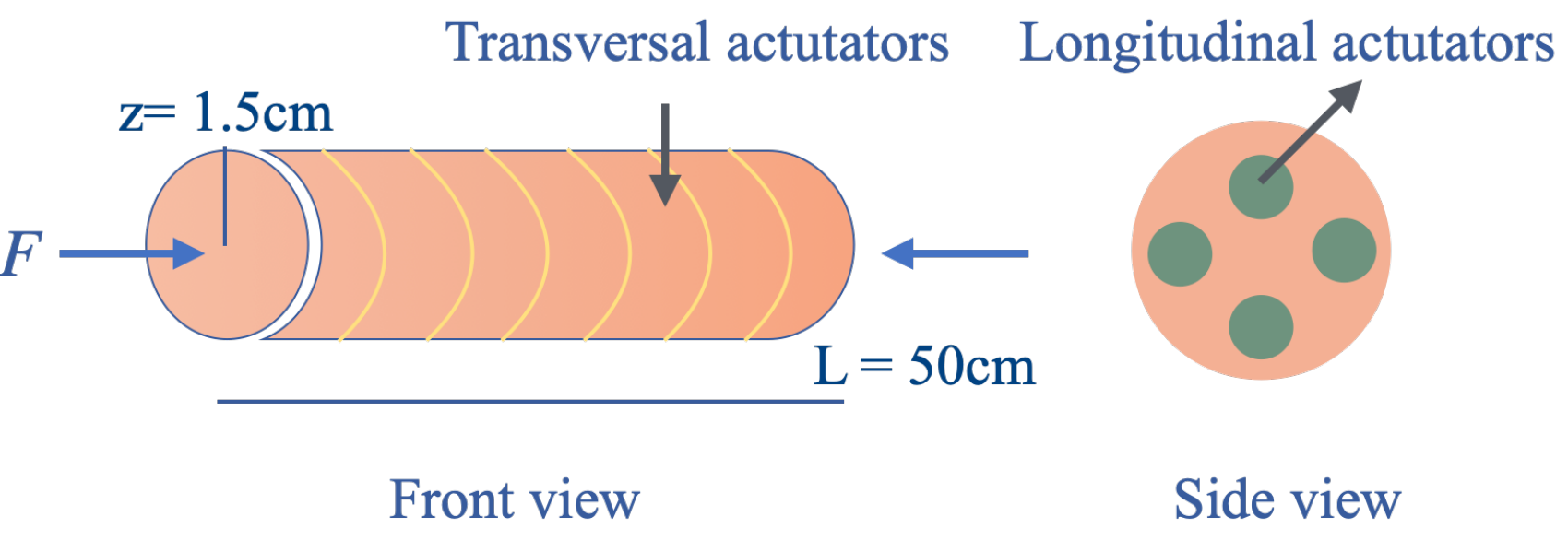}\label{fig:stiffness config}
    }
    \subfigure[]{
    \includegraphics[width=.45\linewidth]{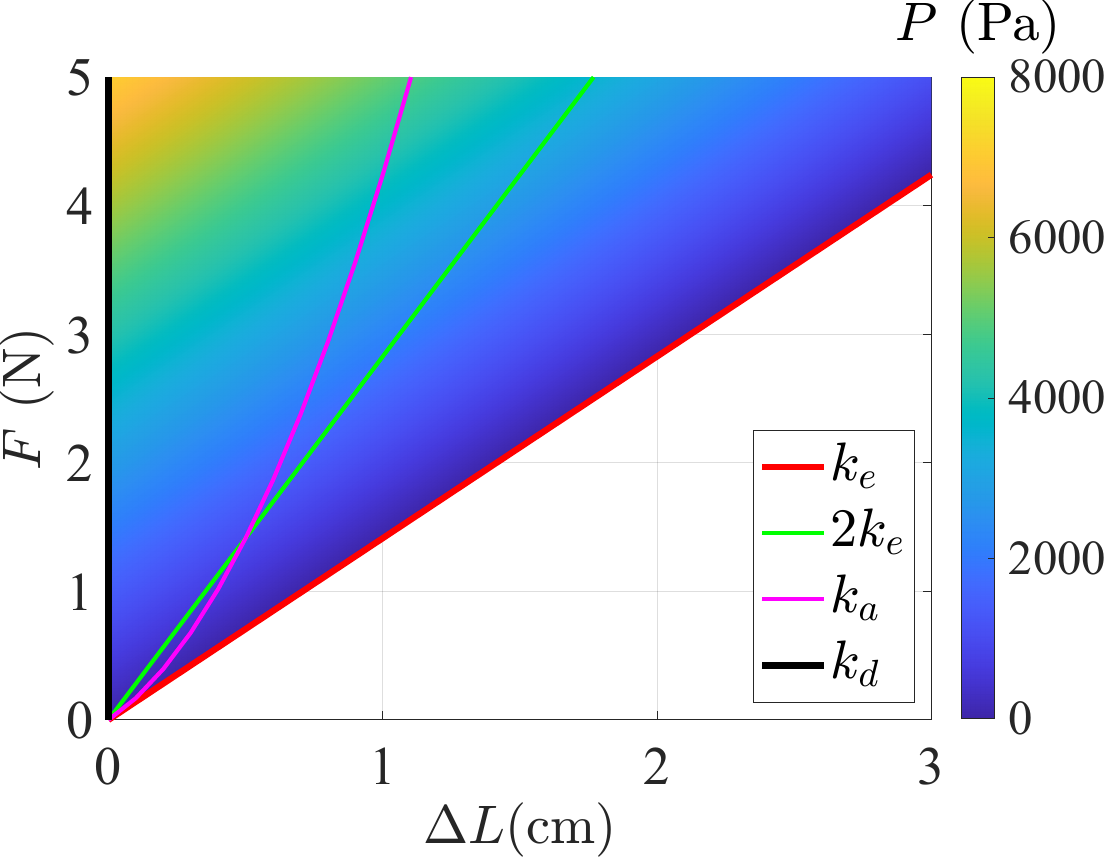}\label{fig:stiffness contour}
    }
    \subfigure[]{
    \includegraphics[width=.45\linewidth]{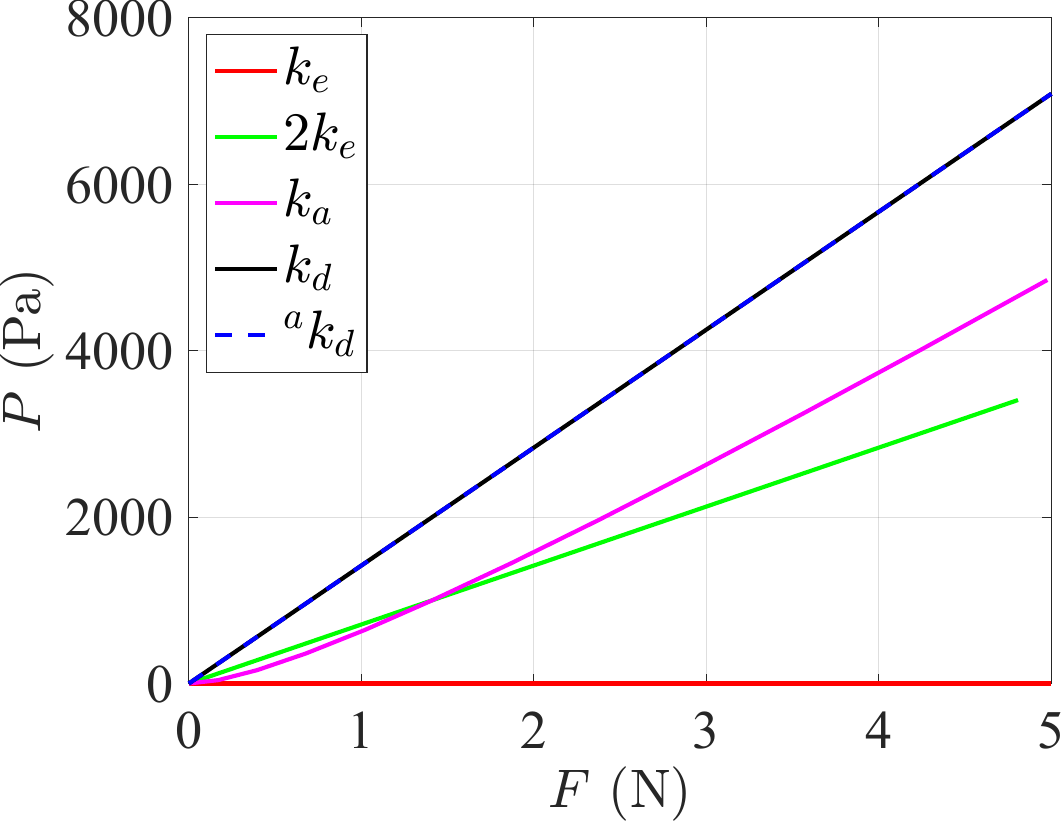}\label{fig:lm vs tm}}
    \caption{Stiffness Tuning of a soft manipulator. (a) The initial configuration of the soft manipulator. The transversal actuator is distributed along the length of the axis, contracting the boundary of the cross-section uniformly. The applied stress  $f_0$ on transversal actuators is consistent along the entire length of the axis. (b) The axial stiffness is determined by $(P, F)$ but is measured by the slope of $F-\Delta L$ diagram. The heatmap represents the active transversal loads to maintain a particular axial stiffness. (c) The combination of axial tension and transversal loads for different axial stiffness.}
    \label{fig:stiffness}
\end{figure}

Stiffness tuning widely exists in soft creature to resist the external force by regulating the stress in the muscles\cite{kier1985tongues}. To evaluate the variable stiffness properties, we conducted a series of tests on the axial direction of the soft manipulator with four parallel longitudinal actuators and a transversal actuator distributed homogeneously along the length of the manipulator, as shown in \figurename~\ref{fig:stiffness config}. The manipulator is defined as a 50-cm length cylinder of 1.5cm diameter, with Young modulus as 0.1MPa and Poisson's ratio as 0.4999. 

Since only tension is allowed in the longitudinal actuators, the soft manipulator can resist the external axial tension by pulling the cables inside the soft body. Such tunable stiffness can be evaluated by the classic Cosserat rod model. However, when there is external axial compression to the soft manipulator, the classic Cosserat rod model fails to capture the axial elasticity caused by the cross-sectional contraction. Here, we employ our extended model to evaluate the internal tension caused by the transversal actuator. We parameterized the axial stretch $\nu_3$ and the inflation ratio $\rho$ by 2nd-order Legendre's polynomial and 1st-order Hermite's polynomial with Neumann boundary condition, respectively. The tests are simplified to only maintain stretch and inflation for the convenience of the computation. We apply external axial compression forces and transverse actuation loads to the manipulator, with forces ranging from 0 to 5 N and loads ranging from 0 to 8 KPa, respectively.

The axial stiffness is defined as $ K = F / \Delta L$, where $F$ is the external axial compression force and $\Delta L $ is the axial displacement. In the absence of internal actuation loads, the passive axial stiffness is given by $K_e = EA / L$, according to the definition of isotropic materials, where $L$ is the original length and $A$ is the original cross-sectional area of the manipulator. When the transversal actuator contracts the cross-section, it generates active internal tension, preventing the axial displacement caused by external compression, which leads to tunable axial stiffness. Denote the stress applied by the transversal actuator as $P$, and for a given pair of $(P, F)$, the corresponding axial displacement $\Delta L$ are obtained by solving \eqname~\ref{eqn: dyn n} and \eqname~\ref{eqn: dyn Q}. The results are presented in \figurename~\ref{fig:stiffness contour}, where the active stiffness is defined as the slope of the $F-\Delta L$ diagram. Therefore, any desirable stiffness greater than $K_e$ can be achieved by applying the appropriate active transverse loads. As transverse loads increase, the internal tension balances the external compression forces, resulting in no axial displacement. Consequently, the manipulator exhibits rigid behavior. The transversal load that achieves rigid behavior is given by $r = \pi z^2 P = -F/\nu^0$. We choose four tunable stiffnesses, $K_e$, $2K_e$, nonlinear stiffnesses $K_a$, and rigid stiffnesses $K_d$ to show the corresponding combination of $(P, F)$ in \figurename~\ref{fig:lm vs tm}. The $K_e$ line in \figurename~\ref{fig:stiffness contour} precisely represents the lower bound of the feasible range of transverse loads. Additionally, the simulated $K_d$ overlaps with the analytical $K_d$ in \figurename~\ref{fig:lm vs tm}, demonstrating the accuracy of our model.

\subsection{Reaching of the Octopus Arm}\label{sec: reaching}
\begin{figure}[tb!]
    \centering
    \includegraphics[width=.8\linewidth]{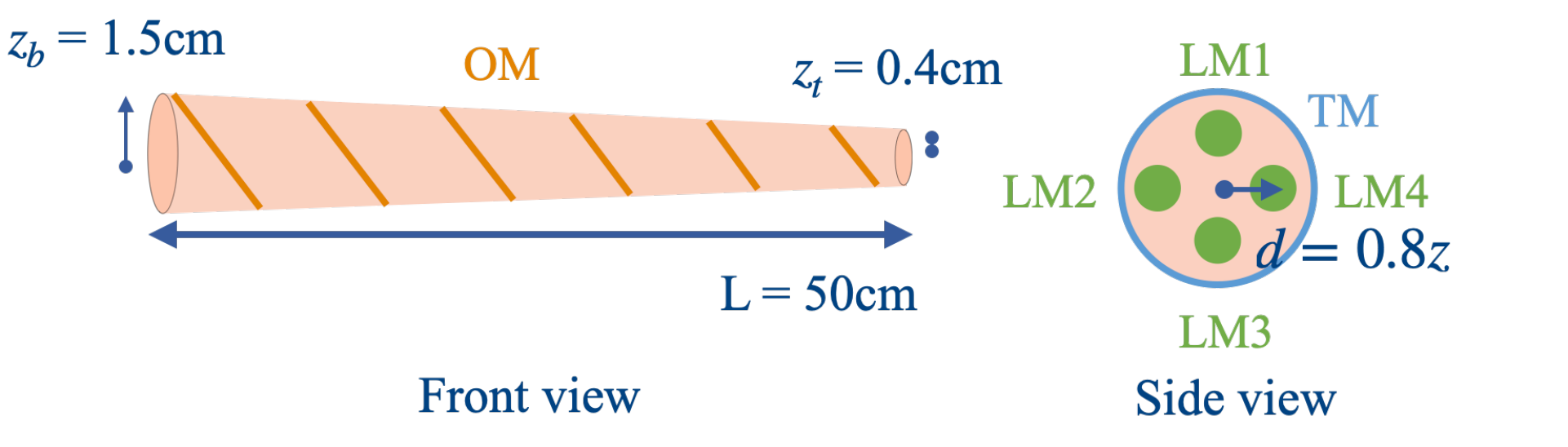}
    \caption{Configuration of the octopus-arm inspired soft manipulator. It consists of a flexible soft body, a group of four longitudinal muscles (LM), one oblique muscles (OM) winding clock-wise along the length, and a set of transversal muscles (TM) on the boundary of all the infinitesimal cross sections.}
    \label{fig:octopus arm}
\end{figure}

The reaching motion of the octopus arm involves both the elongation and bend formed nearby the base, which is then propagated toward the tip \cite{hanassy2015stereotypical}. During the reaching motion, a wave of co-contraction of the transversal and longitudinal muscle passing from base to tip is observed, causing the increase in the flexural stiffness of the arm, \cite{gutfreund1996organization,gutfreund_patterns_1998,yekutieli2005dynamic}. Inspired by the structure of an octopus arm, we model a soft manipulator that mimics its key anatomical arrangement, as shown in \figurename~\ref{fig:octopus arm}. The main body of the soft manipulator is a truncated cone with base radius $z_b = 1.5$ cm, tip radius $z_t = 0.4$ cm, and length $L = 50$cm. The soft manipulator is activated by two types of actuators: a group of four longitudinal muscles (LM) which only allow tension, evenly distributed inside the cross-section from base to tip, each located $0.8 z(s)$ away from the manipulator’s centroid. The second type is the transverse muscles (TM), which are positioned on the boundary of all the infinitesimal cross-sections from base to tip and only allow the contraction on the cross-sections. Unlike the cable-driven soft manipulator, the tension or the stress applied by LM or TM can vary across time and space so that a bending propagation can be achieved.

\begin{table}[tb!]
    \centering
    \caption{Parameters for octopus arm and the environment}
    \label{tab:octopus}
    \begin{tabular}{ccc}
        \hline
        Parameter               & Value & Interpretation \\ \hline
        $\rho_0$                 &   $1e3$ $kg/m^3$    &  density of the soft body             \\
        $E$        &   2 $KPa$    &  Young's modulus             \\
        $\nu^0$         &   0.4999    &  Poisson's ratio             \\
        $\eta$         &  120 $Pa \cdot s$     & shear viscosity              \\
        $\rho_a$    &   $1e3$ $kg/m^3$    &   density of the water            \\
        $B_1$, $B_2$ &   0.6    &    added-mass coefficient           \\
        $C_L$       &   -0.1    &    lift coefficient           \\
        $C_D$       &  1.1     &     drag coefficient          \\ \hline
    \end{tabular}
\end{table}

\begin{figure}[tb!]
    \centering
    \subfigure[]{
        \includegraphics[width=.31\linewidth]{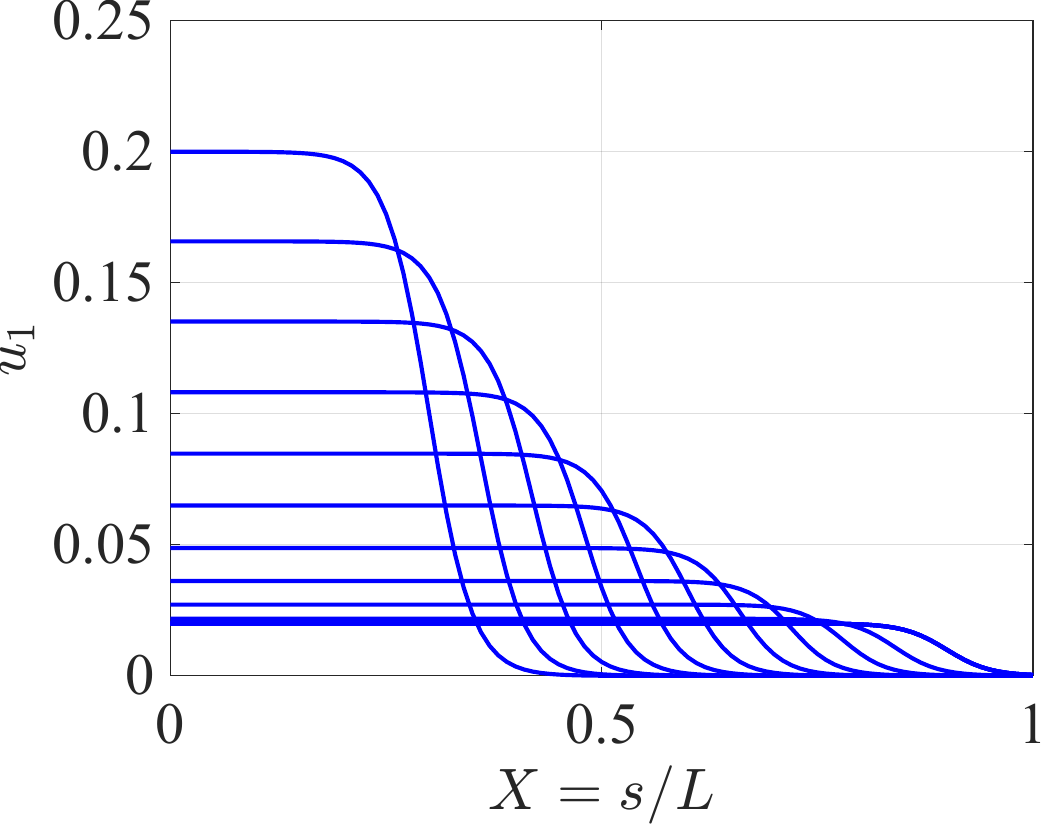}\label{fig:lm release}
    }
    \subfigure[]{
        \includegraphics[width=.31\linewidth]{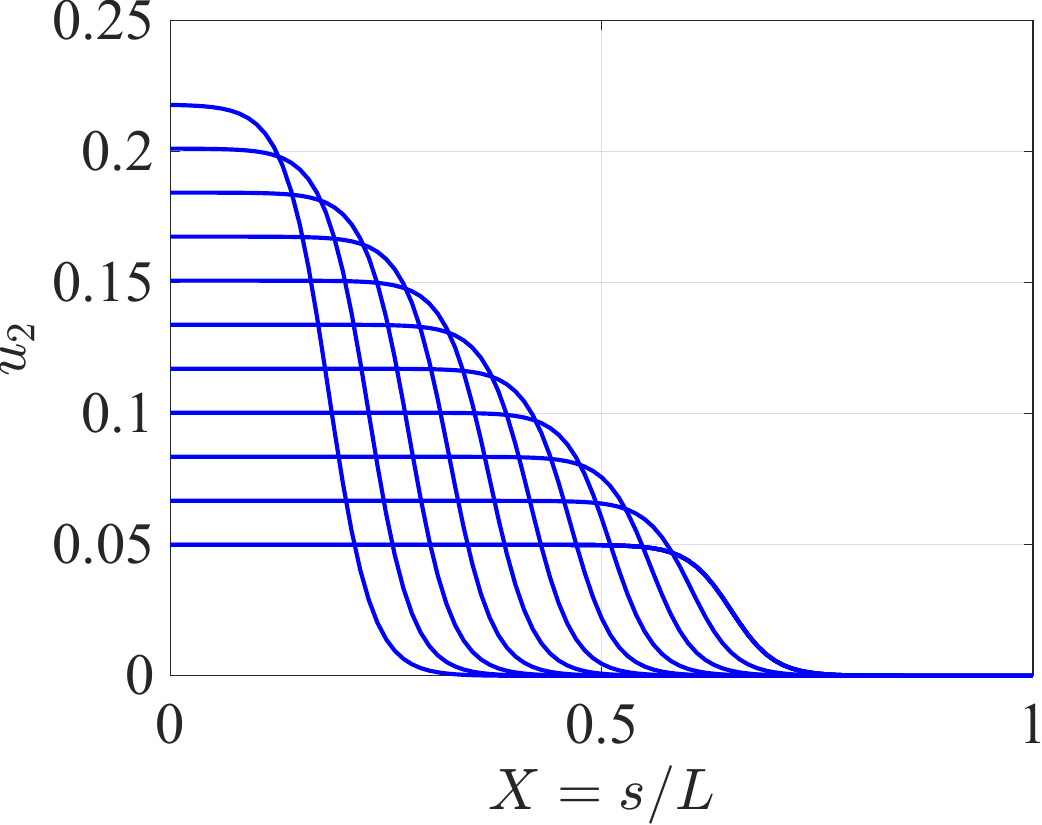}\label{fig:lm contract}
    }
    \subfigure[]{
        \includegraphics[width=.31\linewidth]{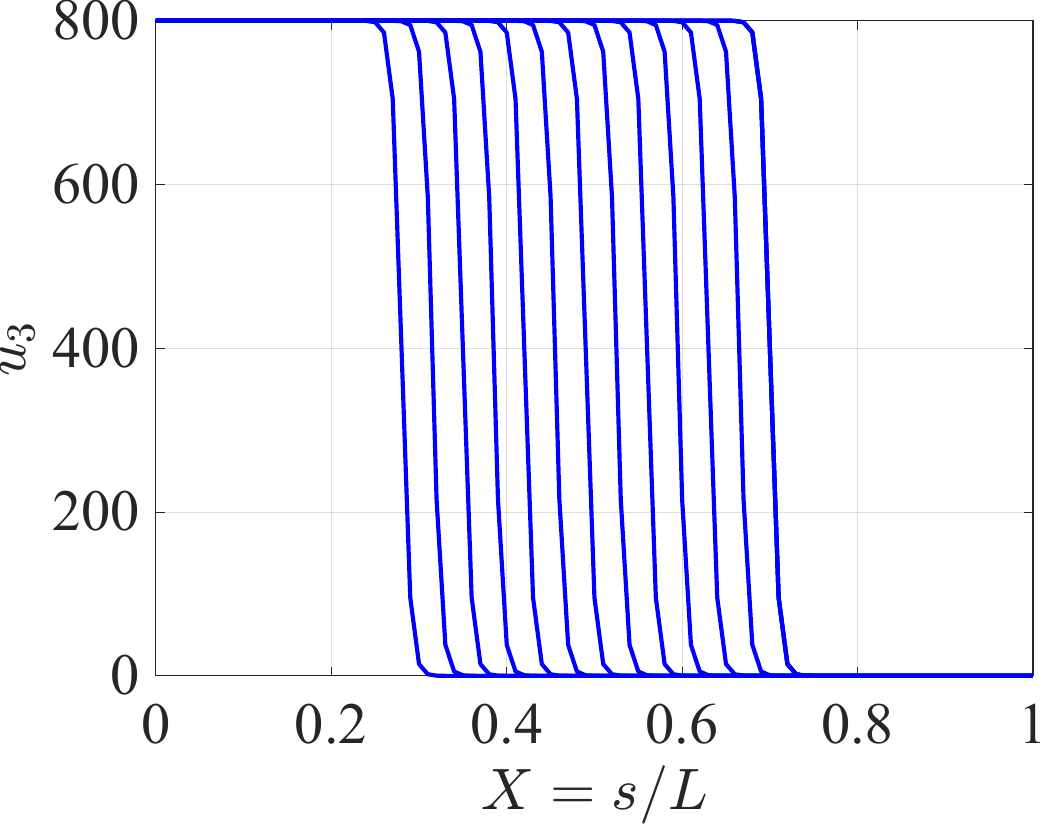}\label{fig:tm contract}
    }
    \caption{Activation traveling wave profiles for all kinds of actuators. Time advances from left to right. (a) Relaxation for aboral LM. (b) Relaxation for oral and side LM. (c) Contraction for TM}
    \label{fig:activated wave}
\end{figure}

To fully capture the flexibility of the octopus arm and the complex underwater environment the octopus arm interacts with, we design the parameters for the soft manipulator and the environment as shown in \tablename~\ref{tab:octopus}. For the convenience of computation, only the planar bending $\kappa_2$, stretch $\nu_3$, and inflation ratio $\rho$ are considered. These parameters are parameterized using 10th-order Legendre polynomials for $\kappa_2$, 4th-order Legendre polynomials for $\nu_3$, and 4th-order Hermite polynomials for $\rho$. Inspired by \cite{wang2022control}, where the pure bending propagation is achieved through the stiffening wave from base to tip on a Kirchhoff rod, we activate all LM and TM by the traveling sigmoid wave following the pattern
\begin{equation}
    u(X,t) = -\alpha(t) \brk{1 - \frac{1}{1+\text{exp}\brk{-\sigma(X - \mu(t))}}}
    \label{eqn:sigmoid}
\end{equation}
where $X = s/L$ denotes the normalized length, $\alpha(t)$ is the magnitude over time, $\mu(t)$ is the position of the traveling wave, and $\sigma$ is the ramping coefficient. The activation wave functions and profiles for all kinds of actuators are listed in \tablename~\ref{tab:wave} and \figurename~\ref{fig:activated wave} respectively.
\begin{table}[tb!]
    \centering
    \caption{Activation functions of reaching motion for all LM, TM}
    \label{tab:wave}
    \begin{tabular}{cccc}
        \hline
        Type & $\alpha(t)$ & $\mu(t)$ & $\sigma$ \\ 
        \hline
        LM 1 $u_1$ & $\text{max}(0.072t^2-0.144t+0.2, 0.02)$& $\text{min}(0.24t+0.3, 0.9)$& 40 \\
        LM 2-4 $u_2$ & $\text{max}(-0.0672t+0.218, 0.05)$ & $\text{min}(0.1868t+0.183, 0.65)$ & 40 \\
        TM $u_3$ & 800 & $\text{min}(0.14t+0.28, 0.7)$ & 200 \\ \hline
    \end{tabular}
\end{table}

\begin{figure}[tb!]
    \centering
    \includegraphics[width=\linewidth]{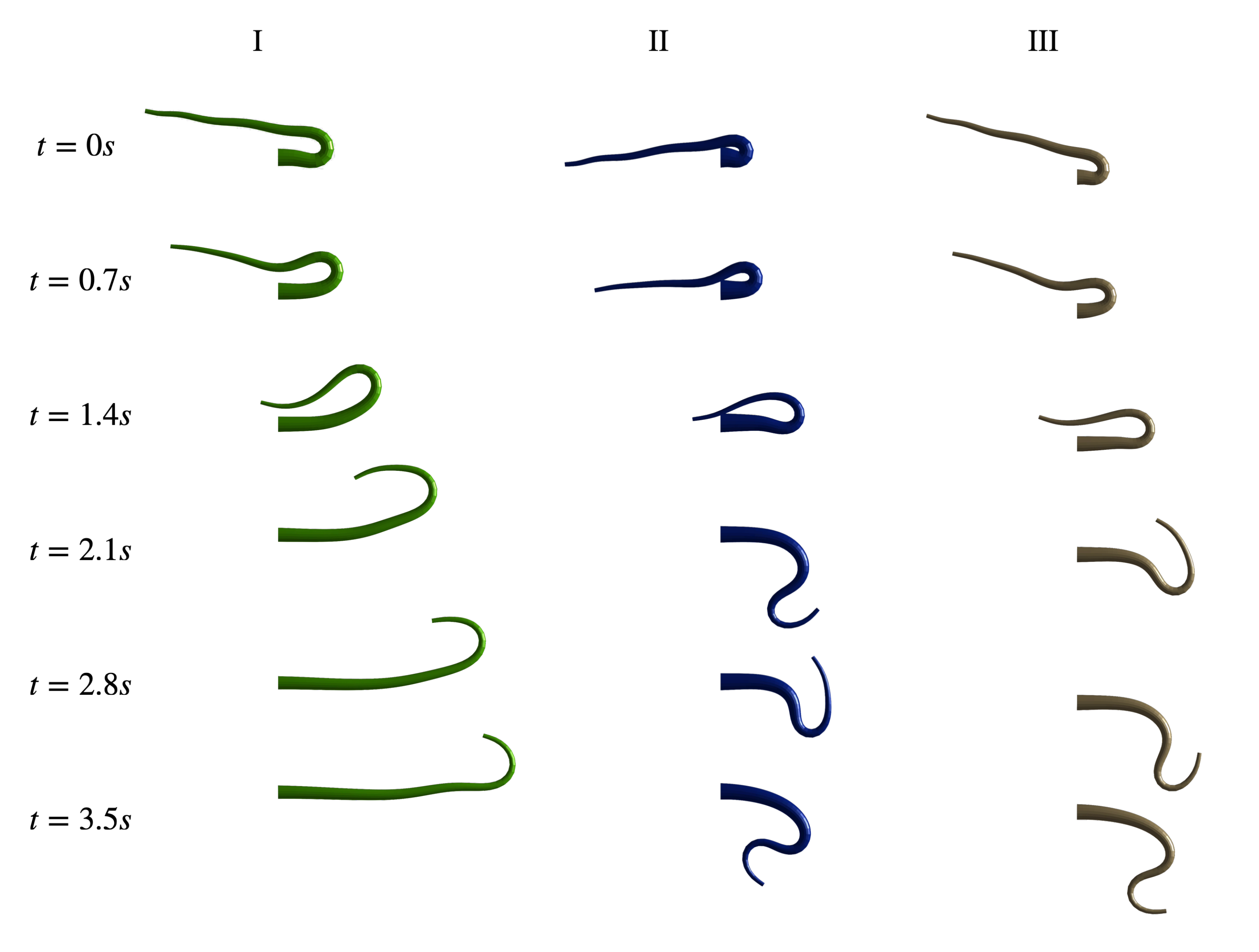}
    \caption{From top to bottom, snapshots of the octopus arm reaching motion over time at 0.7-second intervals. Column I: Extended model with TM contraction. Column II: Extended model without TM contraction. Column III: Classic Cosserat model.}
    \label{fig:snapshots}
\end{figure}

\begin{figure}[tb!]
    \centering
    \subfigure[]{
        \includegraphics[width=.31\linewidth]{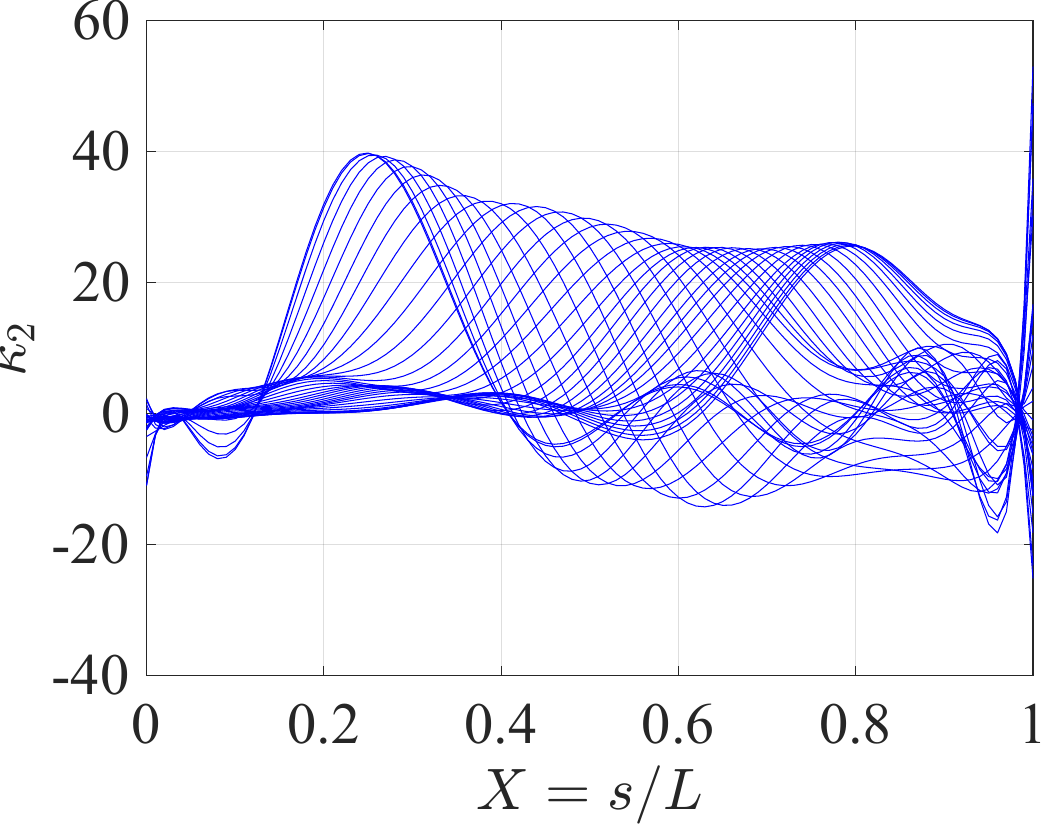}\label{fig:curve over time}
    }
    \subfigure[]{
        \includegraphics[width=.31\linewidth]{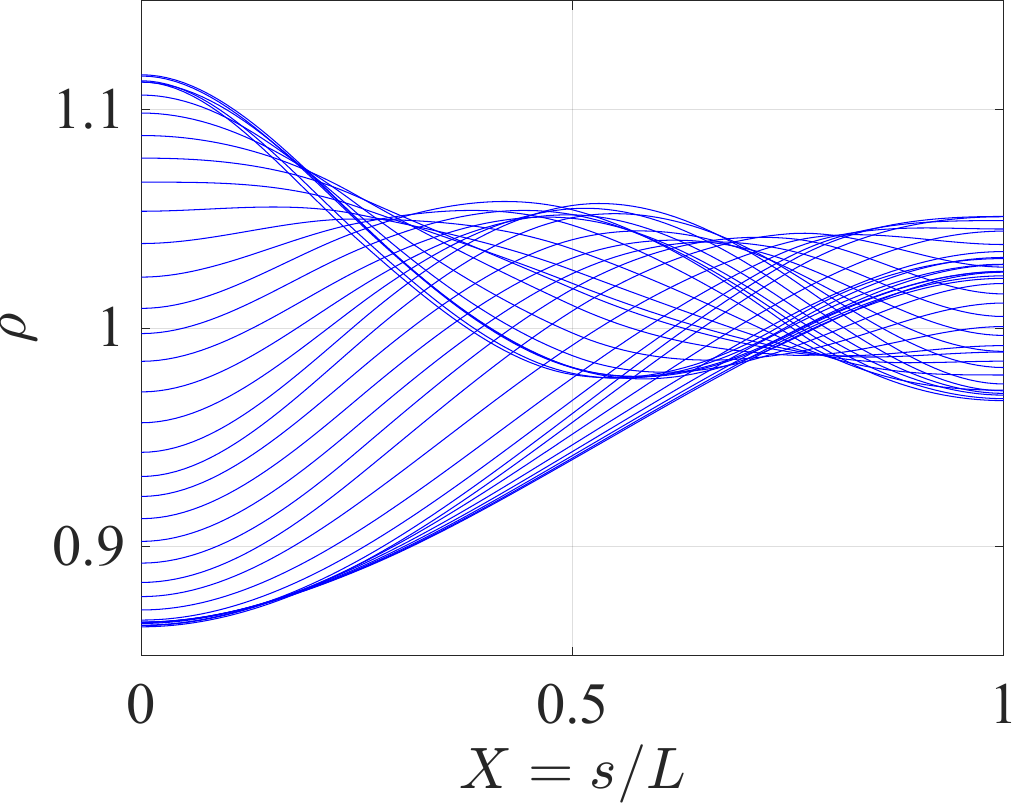}\label{fig:rho over time}
    }
    \subfigure[]{
        \includegraphics[width=.31\linewidth]{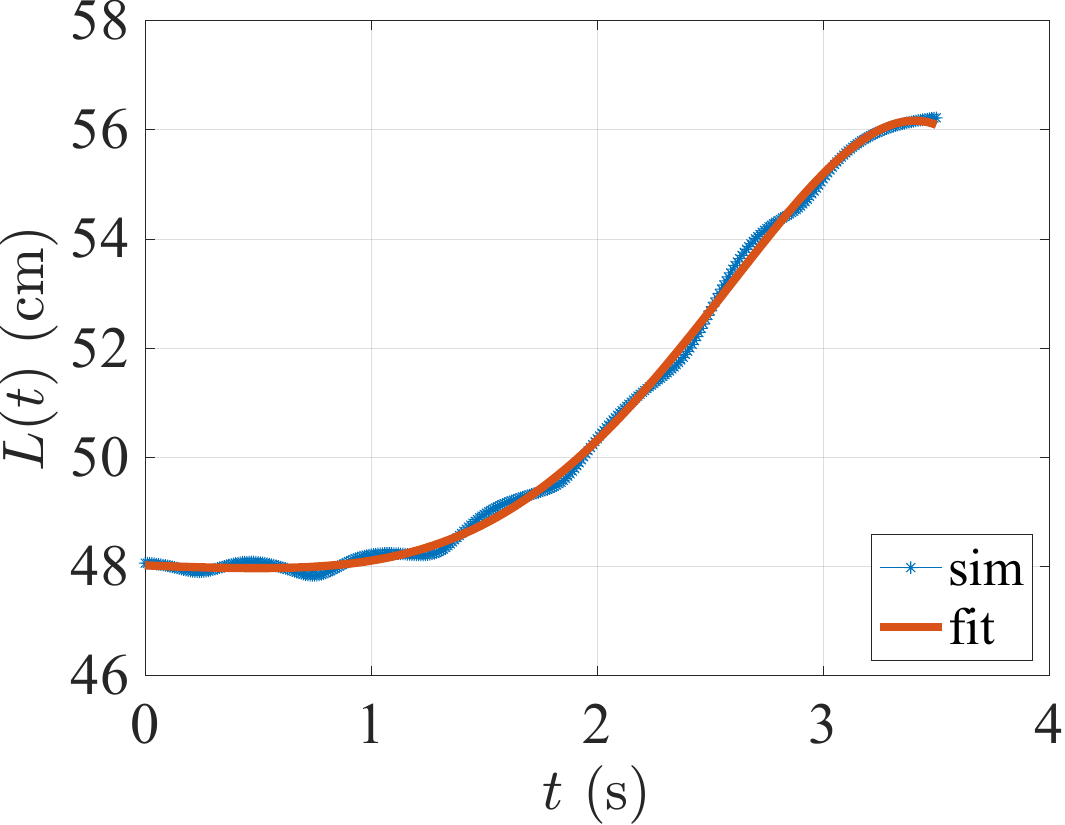}\label{fig:arm length}
    }
    \subfigure[]{
        \includegraphics[width=.31\linewidth]{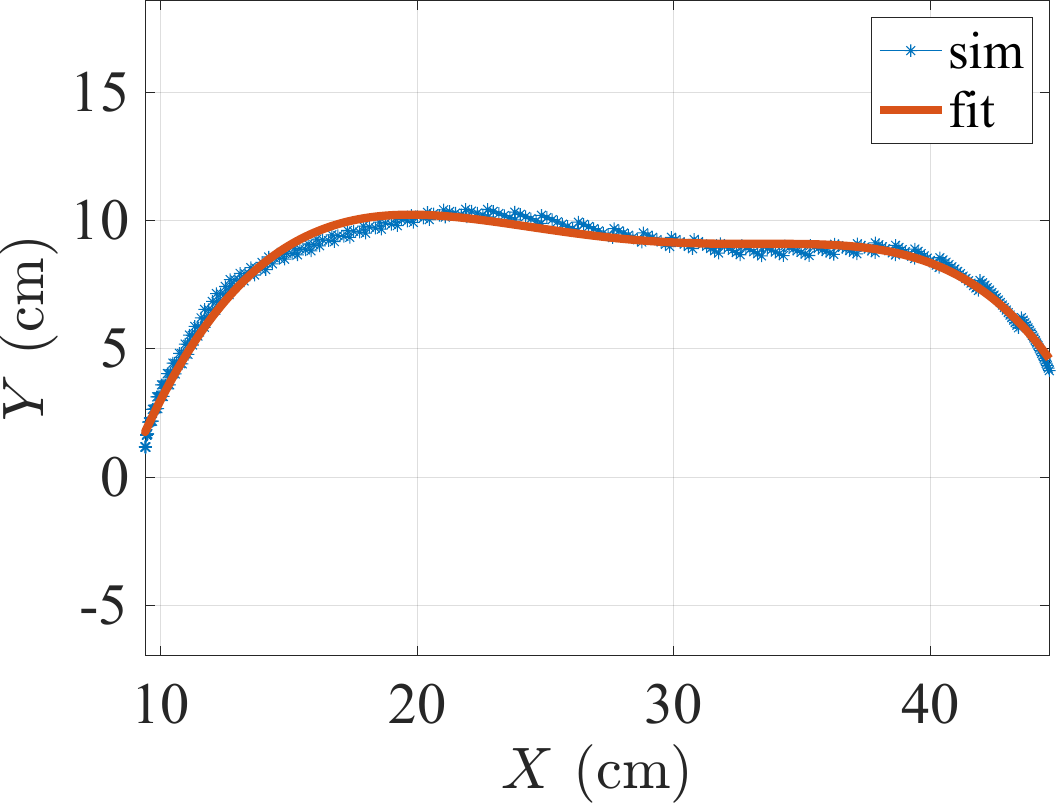}\label{fig:bp pos}
    }
    \subfigure[]{
        \includegraphics[width=.31\linewidth]{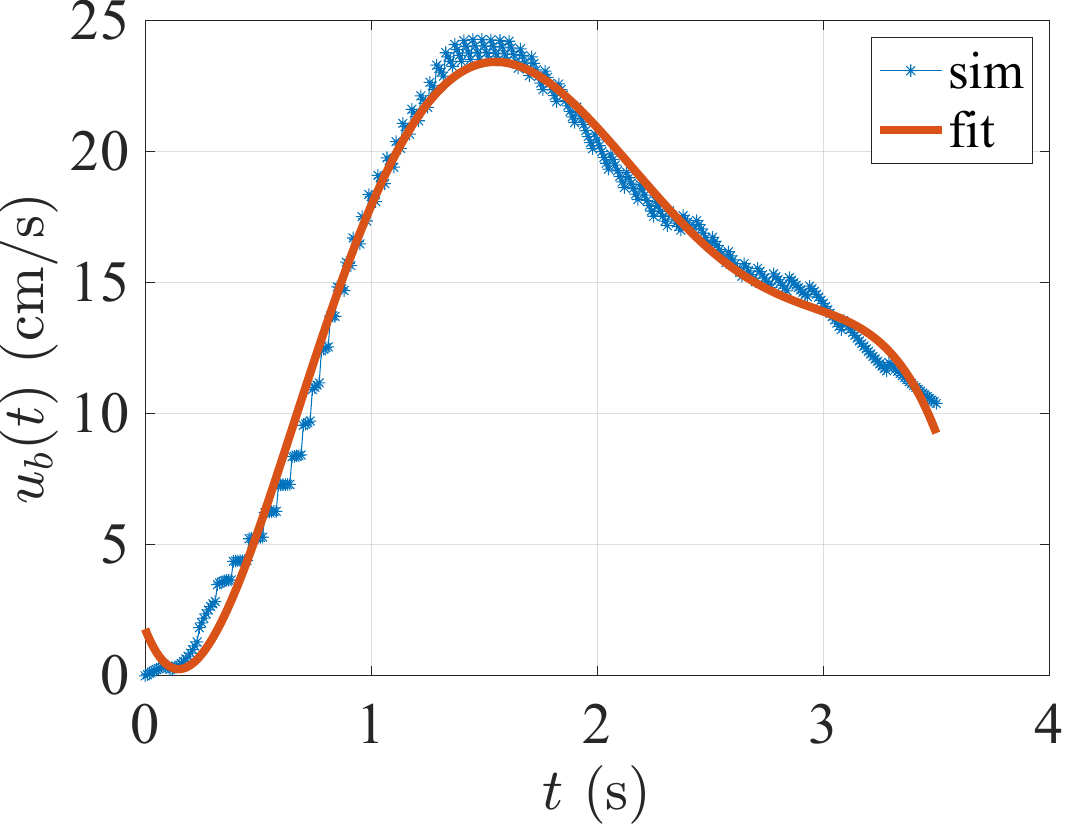}\label{fig:bp velocity}
    }
    \subfigure[]{
        \includegraphics[width=.31\linewidth]{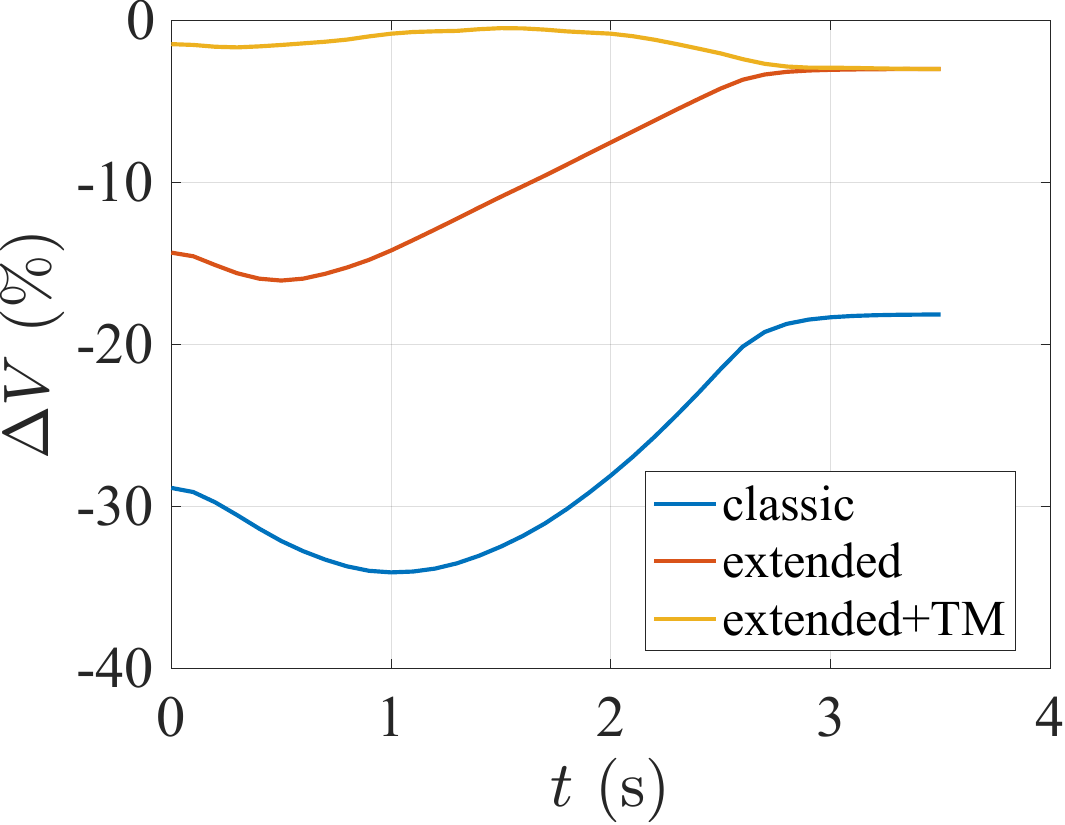}\label{fig:vol change}
    }
    \caption{Biological profiles of the octopus arm. (a) The evolution of the curvature over time. Time advances from left to right (b) The evolution of the inflation ratio over time. Time advances from left to right (c) The arm length over time. (d) The bend point trajectory over time. (e) The bend point velocity over time. (f) The volume change over time.}
    \label{fig: bio profiles}
\end{figure}

The simulation runs on MacBook Pro with Apple Chip 3 and 16GB memory and completes with real time factor 0.45, which is faster than real time. The result of the manipulator's shape over time is shown in \figurename~\ref{fig:snapshots} column 1. We reproduce the octopus-like bending propagation in our simulation\cite{wang2022control}. As shown in \figurename~\ref{fig:curve over time}, the bend point, which is defined as the local maximum curvature point $s_b = \argmax_s \kappa_2(s)$ subjecting to $\kappa_2'(s)=0$, maintains at a relatively large curvature, and travels from nearby base to tip in 3.5 seconds. In addition to pure bending propagation, the octopus arm also exhibits the elongation behavior caused by the contraction of the TM. \figurename~\ref{fig:rho over time} illustrates the evolution of the inflation ratio $\rho$ over time, where the cross-section nearby the base gradually contracts and the tip oscillates about $\rho=1$. To further evaluate the performance of our model, we investigate the arm elongation, bend point trajectory, and bend point velocity over time. The length of the arm is measured from the linear strain along the length of the arm:
\begin{equation*}
    L(t) = \int_0^L \norm{\bnu(s, t)} ds
\end{equation*}
where $\bnu(s, t)$ is given by the linear part of $ \Phi_\xi(s) \bq_\xi(t) + \bq_\xi^*$ at each time step. The bend point trajectory $\br(s_b)$ is given by the coordinates and its velocity is calculated by the magnitude of the linear velocity at $s_b$:
\begin{equation*}
    u_b(t) = \norm{\text{diag}(0, 0, 0, 1, 1, 1)\bJ(\bq_\xi(t), s_{b})\dot{\bq}_\xi(t)}
\end{equation*}
We plot the simulated biological profiles and smooth them with the 5th order polynomial in \figurename~\ref{fig:arm length}, \ref{fig:bp pos}, \ref{fig:bp velocity}. The bend point trajectory travels a total distance of 40.70 cm, while the arm elongates from 48.06 cm to 56.22 cm as the TM contracts from base to tip, accounting for 20.05\% of the traveling distance of the reaching motion. This result indicates that the arm elongation is critical to the reaching motion and is consistent with the biological profiles of the real octopus arm studied in \cite{hanassy2015stereotypical}. The velocity of the bend point starts from 0 and reaches a peak of 24.29 cm/s at 1.49 s, and then gradually decreases to 10.39 cm/s when the bend propagation ends. Its bell shape is typical for reaching movements.

Finally, we compare our model with the extended model without TM contraction and the classic Cosserat model. The results are shown in \figurename~\ref{fig:snapshots} column II, \ref{fig:snapshots} column III, respectively. The classic Cosserat model fails to capture the deformation of the cross section by nature, and hence it is not suitable for studying the complex movements involving TM activations. The extended model without TM contraction also fails to maintain the stretching posture of the arm, which indicates the importance of the TM contraction in the stiffening wave hypothesis proposed by \cite{gutfreund1996organization,gutfreund_patterns_1998}. To further validate the incompressibility of our model, we compute the volume change of the manipulator over time
\begin{equation*}
    \Delta V(t) = \frac{\int_0^L \pi\rho^2z^2\norm{\bnu} ds}{\int_0^L \pi z^2 ds} -  1
\end{equation*}
and plot the results in \figurename~\ref{fig:vol change}. The classic Cosserat model, where the volume preservation is never modeled, shows a significant volume change from -36.01\% to -18.14\%, while the extended model without TM contraction shows a relatively smaller volume change, from -16.04\% to -2.98\%. Our model maintains the volume of the manipulator, with maximum volume change at -2.99\%. The volume conservation is obvious in the extended Cosserat model since a rough incompressibility constraint is accounted for in the constitutive law. However, the volume change is inevitable as the internal actuator loads increase due to the lack of strict volume invariant constraints. The extended model offers an advantage over the classic model by accurately capturing the deformation of the cross-section and incorporating volume preservation, allowing for a more realistic representation of complex movements.

\subsection{Fetching of the Octopus Arm}
In this section, we present the fetching motion of the octopus arm. After reaching the target, the octopus shortens its arm and propagates bending backward to feed the food to its mouth. Twisting occurs when the tip collides with the body \cite{hochner_embodied_2023}. To mimic the fetching motion, we use the same manipulator described in \sectioname~\ref{sec: reaching} and activate the extra oblique muscle (OM) shown in \figurename~\ref{fig:octopus arm} at the collision point. The OM coordinates w.r.t the local frame is defined as
\begin{align*}
    &Y_1 = 0.8\brk{z_b - (z_b-z_t)s/L}\cos{\brk{12\pi s/L}} \\
    &Y_2 = -0.8\brk{z_b - (z_b-z_t)s/L}\sin{\brk{12\pi s/L}}
\end{align*}
so that OM winds as a conical helix along the length of the arm. To showcase the main 3D motion, we parameterize bending $\kappa_1$, $\kappa_2$, and twisting $\kappa_3$ using a fourth-order Legendre polynomial, stretch $\nu_3$ using a second-order Legendre polynomial, and inflation ratio $\rho$ using a second-order Hermite spline. The manipulator is actuated by traveling sigmoid waves (\eqname~\ref{eqn:sigmoid}) applied to all kinds of actuators for the first 8 seconds. The parameters for the activation functions are listed in \tablename~\ref{tab:fetch}, where $H(t)$ denotes the Heaviside function.
\begin{table}[tb!]
    \centering
    \caption{Activation functions of fetching motion for all LM, TM, OM}
    \label{tab:fetch}
    \begin{tabular}{cccc}
        \hline
        Type & $\alpha(t)$ & $\mu(t)$ & $\sigma$ \\ 
        \hline
        LM 1 $u_1$ & $\text{min}(0.06t+0.05, 0.2)$& $\text{max}(-0.12t+0.9, 0.6)$& 200 \\
        LM 2-4 $u_2$ & $\text{min}(-0.048t+0.04, 0.16)$ & $\text{max}(-0.136t+0.86, 0.52)$ & 200 \\
        TM $u_3$ & 1600 & $\text{max}(-0.1667t+0.9, 0.4)$ & 200 \\ 
        OM $u_4$ & $0.1H(t-6)$ & $0.3(t-6)H(t-6)+(2.4-0.3t)H(t-8)$ & 200
        \\\hline
    \end{tabular}
\end{table}

\begin{figure}[tb!]
    \centering
    \subfigure[]{
        \includegraphics[width=.48\linewidth]{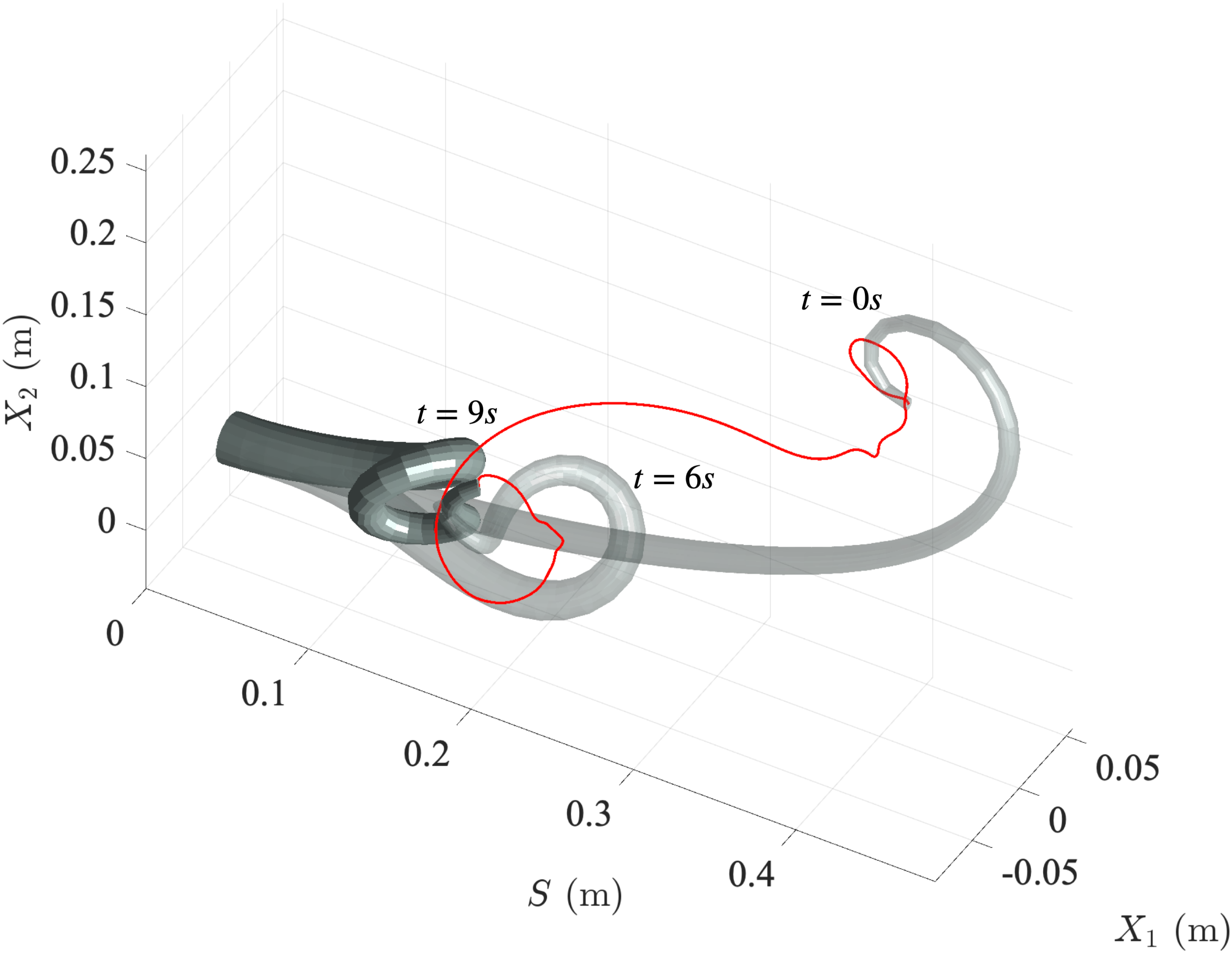}\label{fig:fetching}
    }
    \subfigure[]{\includegraphics[width=.48\linewidth]{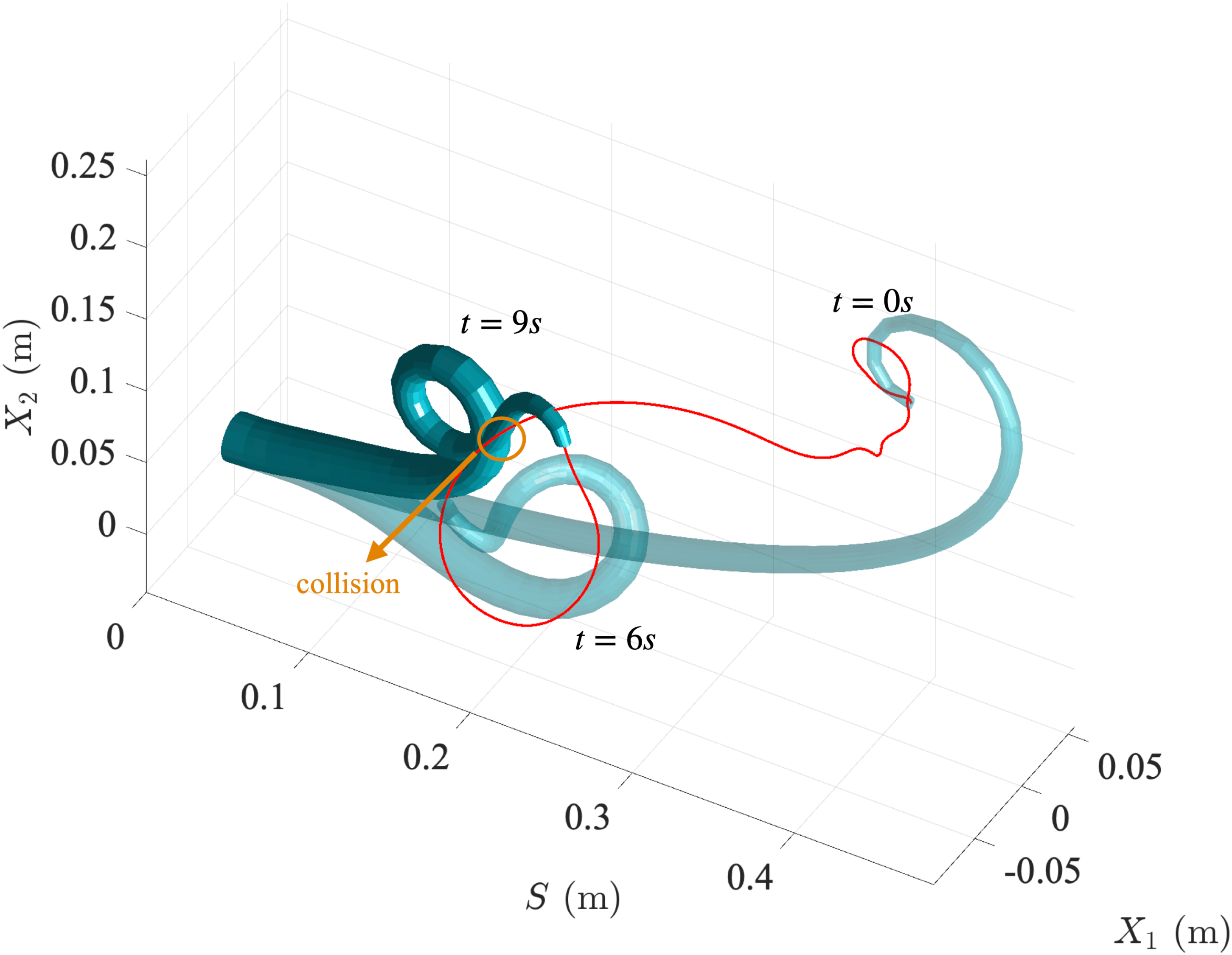}\label{fig:collide}}
    \caption{Snapshots of the 3D motion of the octopus arm. The red curve represents the tip trajectory. (a) Fetching motion. OM actuated to twist at the collision (b) OM not actuated, and as a result, the arm collides with itself.}
    \label{fig:fetch vs collide}
\end{figure}

The simulation results are shown in \figurename~\ref{fig:fetch vs collide}. In \figurename~\ref{fig:fetching}, under the activation of LM and TM,  the octopus arm shortens and rapidly propagates the bending point backward from tip to base. As the bending curvature increases, the tip tends to collide with the body. At $t=6$s, the OM gradually activates from the base to the bending point, avoiding collision and significantly increasing the bending curvature. To compare the effect of the oblique muscle, we simulate the same scenario without the actuated oblique muscle (see \figurename~\ref{fig:collide}). Starting from $t=6$s, the tip gradually collides and cross the arm, a behavior that is unrealistic for a real octopus. The simulation with all muscles activated took 1.74 seconds to complete, whereas the simulation without the oblique muscle took 1.52 seconds.
\section{Conclusion}\label{sec:conclusion}
In this paper,
we develop the extended Cosserat rod model by introducing an extra strain variable, the inflation ratio $\rho$, and derive the constitutive law involving both elasticity and damping from the first principle. We then obtain the equations of motion from the momentum balances and analyze all the possible applied forces on the soft rod. To solve the extended model efficiently, we project the equations of motion from the Euclidean space to the manifold of the parameterized strain configurations using the GVS approach. The paper also includes five appendices, which provide details about the constitutive law, momentum balances, and computation procedures. We also provide open-source \code{MATLAB} codes to guide the reader to implement the extended Cosserat rod model on their applications.

To show the capabilities of the extended Cosserat rod model, we study three applications: the axial stiffness tuning and the reaching and fetching motion of an octopus-arm-like manipulator. In the first application, we demonstrate that the contraction of the inflation ratio $\rho$ can be used to resist axial compression, which is not possible to model in the classical Cosserat rod model. We compare the simulated results with analytical solutions and show that the extended Cosserat rod model can predict the axial stiffness accurately. In the second application, we further validate the stiffening wave hypothesis of the octopus arm reaching by co-contraction of all the LMs and TM. We study the biological profiles of the reaching motion and find them consistent with those of the real octopus. We compare the extended model with transversal actuation, the extended model without transversal actuation, and the classic Cosserat rod model and show the significance of the transversal actuators of the soft manipulator in the complex motions. In the last example, we demonstrate the fetching motion of the octopus arm, which is achieved by the actuation of all types of muscles. The oblique muscles are responsible for the bending and twisting, avoiding collision when the bending curvature is large. 

Despite the fact that the in-plane deformation together with other strains originally considered in the classic Cosserat rod model are effectively simulated, the extended Cosserat rod model is still a simplified model of the soft rod. We do not model the \textit{anistropic} deformation, i.e., the cross-section is not reflective symmetric, or there exists out-of-plane deformation. Even if we roughly impose the incompressibility constraints by relating the diagonal elements of the strain tensor, the model still does not fully preserve volume due to the lack of evaluation of the global volume. The model can be further improved by considering a more general constitutive law, such as the Neo-Hookean hyperelastic model, and incorporating a penalty term for the constant volume in the strain energy function\cite{smith_stable_2018}. For applications, we can further study the control and optimization problems of the soft manipulator, such as the impedance control and motion planning, since our energy-based constitutive law is naturally suitable for creating Lyapunov functions for closed-loop control. Finally, we hope to see more physical soft robotics applications using the extended Cosserat rod model in the future.

\newpage
\appendix
\renewcommand{\thesection}{Appendix \Alph{section}}

\section{Strain Energy Density Rate}\label{sec: strain energy rate}
First, we compute the deformation rate $\Dot{\bF}$ as
$$
\Dot{\bF} = \Dot{\bR}\bR^T\bF + \bR\Dot{\bF}^b = \widetilde{\bW} \bF + \bR\Dot{\bF}^b
$$
where $\widetilde{\bW} = \Dot{\bR}\bR^T \in SO(3)$, $\bF^b = \bD +\bI$. Then we compute the matrix contraction and obtain
$$
\bP : \Dot{\bF} = \tr{(\bP \Dot{\bF}^T)} = \tr \brk{-\bP \bF^T \widetilde{\bW} + \bP\Dot{\bF}^{bT}\bR^T}
$$
Since $\bP\bF^T = \bF \brk{2 \mu \bE + \lambda \tr(\bE)\bI} \bF^T$ is symmetric, then $\tr (-\bP \bF^T \widetilde{\bW}) = 0$, and the line density of the strain energy rate is reduced to
$$
\begin{aligned}
    \Dot{\Phi} &= \int_\Omega \bP : \Dot{\bF} d\Omega = \int_\Omega (\bR \Dot{\bF}^b) : \bP  d\Omega \\
    &= \int_\Omega (\bR \rho\be_\alpha)\cdot \bP \be_\alpha + \brk{\bR \brk{\Dot{\bnu} + X_\alpha \Dot{\rho}' \be_\alpha + X_\alpha\brk{\Dot{\rho}\bkappa + \rho \Dot{\bkappa}}\times\be_\alpha}}\cdot \bP \be_3 d\Omega 
\end{aligned}
$$

Rearrange the above the formula, we obtain \eqname~\ref{eqn: strain rate}.

\section{Material Momentum Balance}\label{sec: traction}
Substituting \eqname~\ref{eqn: dyn 3db},\ref{eqn:q} into \eqname~\ref{eqn: Q prime}, we have
\begin{equation*}
    \begin{aligned}
        Q' &= q - \be_\alpha \cdot \bR^T \int_{\Omega} \bP \be_\alpha d\Omega - \be_\alpha \cdot \bR^T \int_\Omega X_\alpha \brk{\frac{\partial \bP \be_\alpha}{\partial X_\alpha} + \rho_0 \bb-\rho_0 \Ddot{\bp}}d \Omega \\
        &= q - \be_\alpha \cdot \bR^T \brk{\int_\Omega \brk{\bP \be_\alpha + X_\alpha \frac{\partial \bP \be_\alpha}{\partial X_\alpha}}d\Omega + \int_\Omega X_\alpha \rho_0 \bb d\Omega}+ \be_\alpha \cdot \bR^T \brk{\int_\Omega X_\alpha \rho_0 \Ddot{\bp}}d\Omega \\
        &= q - \be_\alpha \cdot \bR^T \brk{\int_\Omega \frac{\partial X_\alpha \bP \be_\alpha}{\partial X_\alpha}d\Omega + \int_\Omega X_\alpha \rho_0 \bb d\Omega}+ \be_\alpha \cdot \bR^T \brk{\int_\Omega X_\alpha \rho_0 \Ddot{\bp}}d\Omega
    \end{aligned}
\end{equation*}

Define
$$
\widetilde{\bP} = \begin{pmatrix}
    \bP\be_1 & | & \bP\be_2 & | & \bm{0}
\end{pmatrix}
$$, then
$$
\nabla \cdot (X_\alpha\widetilde{\bP}) = \frac{\partial X_\alpha \bP \be_\alpha}{\partial X_\alpha}
$$

By divergence theorem, the final expression for $Q'$ is written as
\begin{equation*}
    \begin{aligned}
        Q' &= q - r + \be_\alpha \cdot \bR^T \brk{\int_\Omega X_\alpha \rho_0 \Ddot{\bp}}d\Omega \\
        &= q - \be_\alpha \cdot \bR^T \brk{\oint_C X_\alpha \widetilde{\bP} \cdot \bv dC + \int_\Omega X_\alpha \rho_0 \bb d\Omega}+ \be_\alpha \cdot \bR^T \brk{\int_\Omega X_\alpha \rho_0 \Ddot{\bp}}d\Omega \\
        &= q - \be_\alpha \cdot \bR^T \brk{\oint_C X_\alpha \bP \cdot \bv^b dC + \int_\Omega X_\alpha \rho_0 \bb d\Omega}+ \be_\alpha \cdot \bR^T \brk{\int_\Omega X_\alpha \rho_0 \Ddot{\bp}}d\Omega
    \end{aligned}
    \label{eqn: Q prime final}
\end{equation*}

For statics, we have
\begin{equation*}
    \begin{aligned}
        Q' &= q - r  \\
        &= q - \be_\alpha \cdot \bR^T \brk{\oint_C X_\alpha \bP \cdot \bv^b dC + \int_\Omega X_\alpha \rho_0 \bb d\Omega}
    \end{aligned}
\end{equation*}

\section{Recursive Computation}\label{sec:recursive}
Adjoint operator of $\mathfrak{se}(3)$:
\begin{equation*}
    \text{ad}_{\bxi} = \begin{pmatrix}
        \widetilde{\bkappa} & 0 \\
        \widetilde{\bnu} & \widetilde{\bkappa}
    \end{pmatrix}, \quad
    \text{ad}_{\bbbeta} = \begin{pmatrix}
        \widetilde{\bomega} & 0 \\
        \widetilde{\bu} & \widetilde{\bomega}
    \end{pmatrix}
\end{equation*}
Coadjoint operator of $\mathfrak{se}(3)$:
\begin{equation*}
    \text{ad}^*_{\bxi} = \begin{pmatrix}
        \widetilde{\bkappa} & \widetilde{\bnu}\\
         0 & \widetilde{\bkappa}
    \end{pmatrix}, \quad
    \text{ad}^*_{\bbbeta} = \begin{pmatrix}
        \widetilde{\bomega} & \widetilde{\bu} \\
         0 & \widetilde{\bomega}
    \end{pmatrix}
\end{equation*}
Adjoint map of $SE(3)$
\begin{equation*}
    \text{Ad}_{\bg(s)} = \begin{pmatrix}
        \bR & 0 \\
        \widetilde{\br}\bR & \bR
    \end{pmatrix}
\end{equation*}
Coadjoint map of $SE(3)$
\begin{equation*}
    \text{Ad}^*_{\bg(s)} = \begin{pmatrix}
        \bR & \widetilde{\br}\bR \\
        0 & \bR
    \end{pmatrix}
\end{equation*}
Exponential map of $SE(3)$
\begin{equation*}
    \text{exp}\brk{\widehat{\bOmega}} = \bI_4 + \widehat{\bOmega} + \frac{1}{\theta^2}\brk{1-\cos{\theta}}\widehat{\bOmega}^2+ \frac{1}{\theta^3}(\theta - \sin (\theta))\widehat{\bOmega}^3
\end{equation*}
where $\theta = \norm{\bkappa}$.
Tangent operator:
\begin{align*}
    T_{\bOmega} = 
    &\bI_6 + \frac{1}{2\theta^2}\brk{4 - 4\cos{\theta} - \theta\sin{\theta}}
    \text{ad}_{\bOmega}\\
    &+\frac{1}{2\theta^3}(4\theta - 5\sin{\theta}+\theta\cos{\theta})\text{ad}^2_{\bOmega}\\
    &+\frac{1}{2\theta^4}(2-2\cos{\theta}-\theta\sin{\theta})\text{ad}^3_{\bOmega}\\
    &+\frac{1}{2\theta^5}(2\theta-3\sin{\theta}+\theta\cos{\theta})\text{ad}^4_{\bOmega}
\end{align*}
Fourth-order Zannah quadrature approximation of Magnus expansion
\begin{equation*}
    \bOmega(h) = \frac{h}{2}(\bxi_{1}+\bxi_{2})+\frac{\sqrt{3}h^2}{12}\text{ad}_{\bxi_{1}}\bxi_{2}
\end{equation*}
where subscript $1$ and $2$ represents the Zannah point evaluated at $s+(1/2\mp \sqrt{3}/6)h$

\section{Coefficients of General Dynamics Equation}\label{sec: coeff}
Parameter matrices for \eqname~\ref{eqn: 1st gen eom}:
\begin{align*}
    &\bM_\xi(\bq_\xi, \bq_\rho) = \int_0^L \bJ^T \Bar{\bm{\calM}} \bJ ds \\
    &\bC(\bq_\xi, \bq_\rho, \Dot{\bq}_\xi, \Dot{\bq}_\rho) = \int_0^L \bJ^T \brk{\Bar{\bm{\calM}}\bJ + \Dot{\Bar{\bm{\calM}}} \bJ + \text{ad}_{\bbbeta}\Bar{\bm{\calM}}\bJ} ds \\
    &\bD_\xi = \int_0^L \bPhi_\xi^T \bm{\Upsilon} \bPhi_\xi ds \\
    &\bK_\xi = \int_0^L \bPhi_\xi^T \bm{\Sigma} \bPhi_\xi ds \\
    &\Bar{\bK}_\xi = \int_0^L \bPhi_\xi^T \bm{\sigma} \bPhi_\rho ds \\
    &\bB_\xi(\bq_\xi, \bq_\rho) = \int_0^L \bPhi_\xi^T \bPhi_a ds \\
    &\bF_\xi(\bq_\xi, \Dot{\bq}_\xi, t) = \int_0^L \bJ^T \Bar{\bm{\calF}}
\end{align*}
Parameter matrices for \eqname~\ref{eqn: 2nd gen eom} (assuming $\rho^* \equiv 1$)
\begin{align*}
    &\bM_\rho = \int_0^L \bPhi_\rho^T \rho_0 I_{\alpha \alpha} \bPhi_\rho \\
    &\bD_\rho = \int_0^L \bPhi_\rho^{'T} \eta I_{33} \bPhi_\rho'\\
    &\bK_\rho(\bq_\xi, \Dot{\bq}_\xi) = \int_0^L \bPhi_\rho^{'T} \mu I_{33} \bPhi_\rho' + \bPhi_\rho^T 4\mu A_0 \bPhi_\rho - \bPhi_\rho^T \rho_0 I_{\alpha\alpha}\widehat{\bomega}_{\alpha\alpha}^2 ds\\
    &\Bar{\bK}_\rho = \int_0^L \bPhi_\rho^T\bm{\Sigma}_\rho \bPhi_\xi ds\\
    &\bB_\rho(\bq_\rho) = \int_0^L \bPhi_\rho \bPhi_r ds \\
    &\bF_\rho(\bq_\xi, \Dot{\bq}_\rho) = \int_0^L \bPhi_\rho^T \rho_0 I_{\alpha\alpha}\widehat{\bomega}_{\alpha\alpha}^2 \rho^* ds
\end{align*}

\section{Computation}\label{sec:computation}
\begin{figure}[tb!]
    \centering
    \subfigure[]{\includegraphics[width=.49\linewidth]{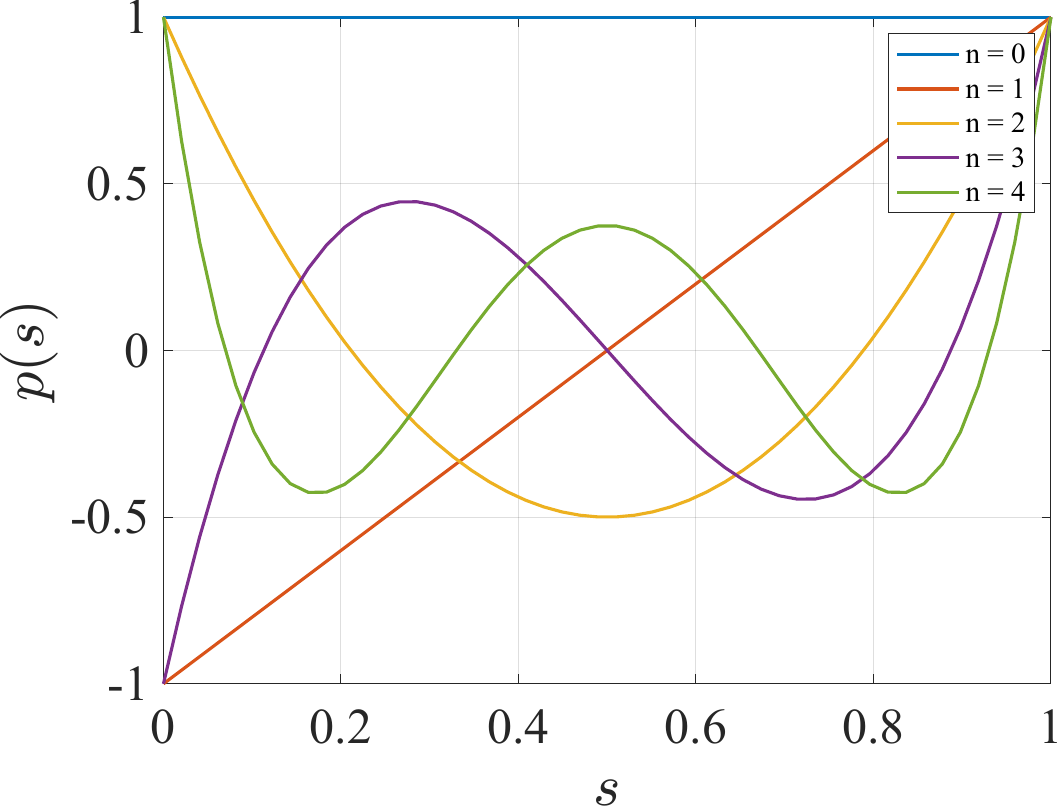}\label{fig:basis legendre}}
    \subfigure[]{\includegraphics[width=.49\linewidth]{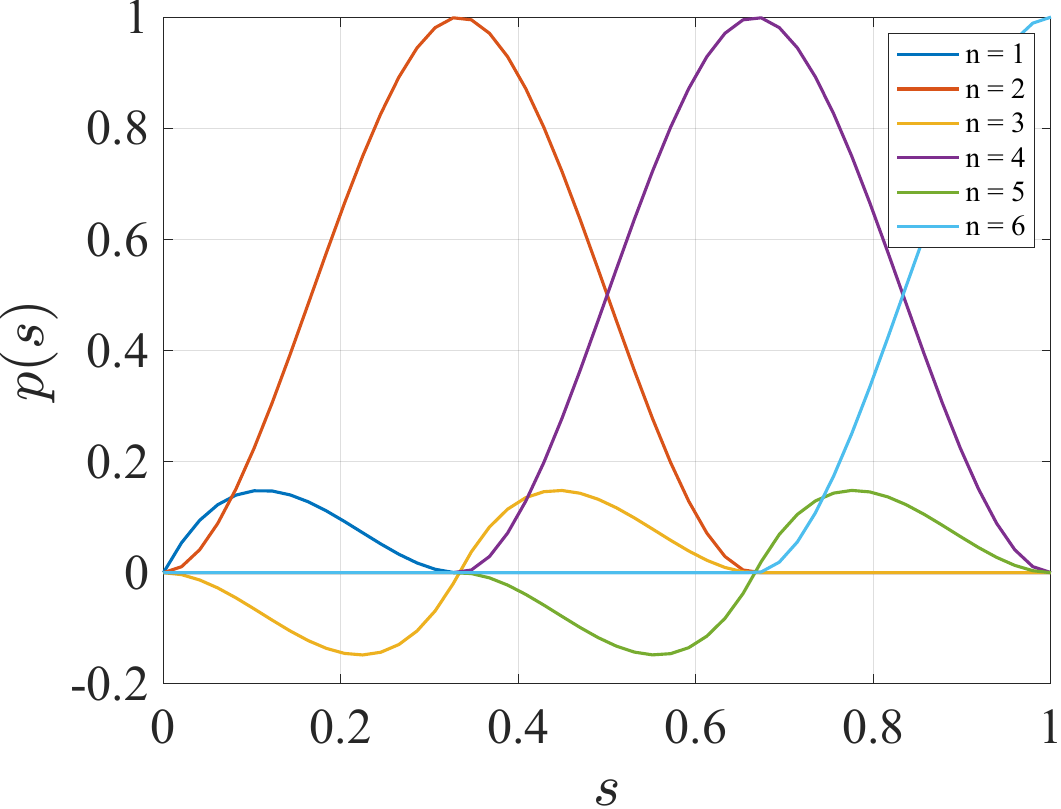}\label{fig:basis hermite}}
    \caption{Basis functions to parameterize twist $\bxi$ and inflation $\rho$. $n$ represents the order of the polynomial (a) Legendre polynomial basis (b) local Hermite basis.}
    \label{fig:bases}
\end{figure}

The strain variables are parameterized by the independent bases so that all the strain variables and their time derivatives are the linear functions of general coordinates. We choose Legendre polynomial as the basis of the strain twist $\bxi$. The standard Legendre polynomial of degree $m$ is given by $P_m(X)$ over the interval $X\in [-1, 1]$, and satisfies the orthogonal condition: $\int_{-1}^{1} P_m(X) P_n(X) dX = 0$ if $m \neq n$, and $P_m(1) = 1$. Each row of the basis is constructed by the $m$-th order Legendre polynomial $P_m(s)$ scaling to the interval $[0, L]$. \figurename~\ref{fig:basis legendre} shows the Legendre polynomial of degree 4. If we model the soft rod with a quadratic bending about $\bd_1$, $\bd_2$,  a quadratic twisting about $\bd_3$, a constant shear about $\bd_1$, $\bd_2$, and a linear stretch about $\bd_3$, so that the basis $\bPhi_\xi$ is written as
\setcounter{MaxMatrixCols}{20}
\begin{equation*}
\bPhi_\xi = \begin{pmatrix}
    P_0 &P_1 &P_2 &0   &0   &0   &0   &0   &0   &0   &0   &0   &0\\
    0   &0   &0   &P_0 &P_1 &P_2 &0   &0   &0   &0   &0   &0   &0\\ 
    0   &0   &0   &0   &0   &0   &P_0 &P_1 &P_2 &0   &0   &0   &0\\
    0   &0   &0   &0   &0   &0   &0   &0   &0   &P_0 &0   &0   &0\\
    0   &0   &0   &0   &0   &0   &0   &0   &0   &0   &P_0 &0   &0\\
    0   &0   &0   &0   &0   &0   &0   &0   &0   &0   &0   &P_0 &P_1
\end{pmatrix} \in \setR^{6\times13}
\end{equation*}

The inflation ratio, on the other hand, is parameterized by the local Hermite spline to meet the boundary conditions stated in \sectioname~\ref{sec: general eom}. Let $\Bar{s} = (s-a)/(b-a)$, $s\in [a, b] \subseteq [0, L]$, then the Hermite spline with mixed boundary condition is defined as
\begin{equation*}
    \begin{aligned}
        p_1(\Bar{s}) = \begin{cases}
            0, & \text{if } a=0\\
            1-3\Bar{s}^2 + 2\Bar{s}^3, & \text{if } a>0
        \end{cases}, \quad &p_2(\Bar{s}) = \Bar{s}-2\Bar{s}^2+\Bar{s}^3 \\
        p_3(\Bar{s}) = 3\Bar{s}^2-2\Bar{s}^3, \quad &p_4(\Bar{s}) = \begin{cases}
            -\Bar{s}^2 + \Bar{s}^3, & \text{if } b<L \\
            0, & \text{if } b=L
        \end{cases}
    \end{aligned}
\end{equation*}
Notice that if $a>0$ and $b<1$, we have $p_k(0) = p_{k+2}(1)$, $p_{k}'(0)=p_{k+2}'(1)$ ($k = 1, 2$), and thus, we can consider $p_k$ and $p_{k+2}$ as one continuous spline over $[0, L]$. In this case, $\bPhi_\rho \in \setR^{1\times 2m}$ with $m$ being the number of pieces of the soft rod.  For example, \figurename~\ref{fig:basis hermite} shows the mixed Hermite spline for a three-pieces soft rod.

The estimation of the pose of the rod requires the integration along the length of the rod. Here, we use the Gaussian-Legendre quadrature \cite{golub1969calculation} to numerically compute the integrals by weighted summing the function values at the discrete points:
\begin{equation*}
    \int_0^L f(s) ds = \sum_{k=1}^{n_i} w_k f(s_k)
\end{equation*}
where $n_i$ is the number of integration points, $w_k$ is the weight of the discrete point $s_k$. The discrete points are computed for both quadrature points and the Zannah collocations to calculate the Magnus expansion of strain twist $\bxi$.

Given the bases $\bPhi_\xi$ and $\bPhi_\rho$, all constant parameterized matrices are derived as specified in \eqname~\ref{eqn: 1st gen eom} and \ref{eqn: 2nd gen eom}. Initially, at $t=0$, the states $(\bq_\xi, \bq_\rho, \Dot{\bq}_\xi, \Dot{\bq}_\rho)$ are provided, allowing for the computation of $\bxi$, $\rho$, $\bg$, $\bJ$, and $\bbbeta$ in the domain $[0, L]$. The parameter matrices for \eqname~\ref{eqn: 1st gen eom} and \ref{eqn: 2nd gen eom} are then recursively calculated along the rod's length using the Gaussian-Legendre quadrature method. For instance, the general mass matrix $\bM_\xi$ is approximated by:
\begin{equation*}
\bM_\xi(\bq_\xi, \bq_\rho) = \int_0^L \bJ^T \Bar{\bm{\calM}} \bJ , ds = \sum_{k=1}^{n_i} w_k \bJ(s_k)^T \bm{\calM}(s_k) \bJ(s_k)
\end{equation*}
With these matrices computed, the dynamical system described in \eqname~\ref{eqn: 1st gen eom} and \ref{eqn: 2nd gen eom} is converted into two sets of second-order ODEs. These are solved using the \code{ode15s} solver in \code{MATLAB} to estimate the state $(\bq_\xi, \bq_\rho, \Dot{\bq}_\xi, \Dot{\bq}_\rho)$ at the next time step. For static problems, an initial guess of the equilibrium state is used. The parameter matrices are updated iteratively, and the error between both sides of \eqname~\ref{eqn: 1st statics} and \ref{eqn: 2nd statics} is computed. The solution of $(\bq_\xi, \bq_\rho)$ at equilibrium is found using the \code{fsolve} method until the convergence of the error is achieved.

\bibliography{ref}

\begin{thebibliography}{10}

\bibitem{mengaldo_concise_2022}
G.~Mengaldo, F.~Renda, S.~L. Brunton, M.~Bächer, M.~Calisti, C.~Duriez, G.~S. Chirikjian, and C.~Laschi, ``A concise guide to modelling the physics of embodied intelligence in soft robotics,'' {\em Nature Reviews Physics}, vol.~4, pp.~595--610, Aug. 2022.

\bibitem{pfeifer2007self}
R.~Pfeifer, M.~Lungarella, and F.~Iida, ``Self-organization, embodiment, and biologically inspired robotics,'' {\em science}, vol.~318, no.~5853, pp.~1088--1093, 2007.

\bibitem{kier1985tongues}
W.~M. Kier and K.~K. Smith, ``Tongues, tentacles and trunks: the biomechanics of movement in muscular-hydrostats,'' {\em Zoological journal of the Linnean Society}, vol.~83, no.~4, pp.~307--324, 1985.

\bibitem{smith1989trunks}
K.~K. Smith and W.~M. Kier, ``Trunks, tongues, and tentacles: moving with skeletons of muscle,'' {\em American Scientist}, vol.~77, no.~1, pp.~28--35, 1989.

\bibitem{chirikjian_hyper-redundant_1994}
G.~S. Chirikjian, ``Hyper-redundant manipulator dynamics: a continuum approximation,'' {\em Advanced Robotics}, vol.~9, pp.~217--243, Jan. 1994.

\bibitem{calisti_octopus-bioinspired_2011}
M.~Calisti, M.~Giorelli, G.~Levy, B.~Mazzolai, B.~Hochner, C.~Laschi, and P.~Dario, ``An octopus-bioinspired solution to movement and manipulation for soft robots,'' {\em Bioinspiration \& Biomimetics}, vol.~6, p.~036002, Sept. 2011.

\bibitem{laschi_soft_2012}
C.~Laschi, M.~Cianchetti, B.~Mazzolai, L.~Margheri, M.~Follador, and P.~Dario, ``Soft {Robot} {Arm} {Inspired} by the {Octopus},'' {\em Advanced Robotics}, vol.~26, pp.~709--727, Jan. 2012.

\bibitem{10.5555/1036292}
J.~M. Selig, {\em Geometric Fundamentals of Robotics (Monographs in Computer Science)}.
\newblock SpringerVerlag, 2004.

\bibitem{brady1982robot}
M.~Brady, {\em Robot motion: Planning and control}.
\newblock MIT press, 1982.

\bibitem{armanini_soft_2023}
C.~Armanini, F.~Boyer, A.~T. Mathew, C.~Duriez, and F.~Renda, ``Soft {Robots} {Modeling}: {A} {Structured} {Overview},'' {\em IEEE Transactions on Robotics}, vol.~39, pp.~1728--1748, June 2023.

\bibitem{10136424}
C.~Della~Santina, C.~Duriez, and D.~Rus, ``Model-based control of soft robots: A survey of the state of the art and open challenges,'' {\em IEEE Control Systems Magazine}, vol.~43, no.~3, pp.~30--65, 2023.

\bibitem{lai2022constrained}
J.~Lai, B.~Lu, Q.~Zhao, and H.~K. Chu, ``Constrained motion planning of a cable-driven soft robot with compressible curvature modeling,'' {\em IEEE robotics and automation letters}, vol.~7, no.~2, pp.~4813--4820, 2022.

\bibitem{webster_design_2010}
R.~J. Webster and B.~A. Jones, ``Design and {Kinematic} {Modeling} of {Constant} {Curvature} {Continuum} {Robots}: {A} {Review},'' {\em The International Journal of Robotics Research}, vol.~29, pp.~1661--1683, Nov. 2010.

\bibitem{jones2006kinematics}
B.~A. Jones and I.~D. Walker, ``Kinematics for multisection continuum robots,'' {\em IEEE Transactions on Robotics}, vol.~22, no.~1, pp.~43--55, 2006.

\bibitem{camarillo2008mechanics}
D.~B. Camarillo, C.~F. Milne, C.~R. Carlson, M.~R. Zinn, and J.~K. Salisbury, ``Mechanics modeling of tendon-driven continuum manipulators,'' {\em IEEE transactions on robotics}, vol.~24, no.~6, pp.~1262--1273, 2008.

\bibitem{camarillo_configuration_2009}
D.~Camarillo, C.~Carlson, and J.~Salisbury, ``Configuration {Tracking} for {Continuum} {Manipulators} {With} {Coupled} {Tendon} {Drive},'' {\em IEEE Transactions on Robotics}, vol.~25, pp.~798--808, Aug. 2009.

\bibitem{chirikjian_closed-form_1995}
G.~S. Chirikjian, ``Closed-{Form} {Primitives} for {Generating} {Locally} {Volume} {Preserving} {Deformations},'' {\em Journal of Mechanical Design}, vol.~117, pp.~347--354, Sept. 1995.

\bibitem{zheng2012dynamic}
T.~Zheng, D.~T. Branson, R.~Kang, M.~Cianchetti, E.~Guglielmino, M.~Follador, G.~A. Medrano-Cerda, I.~S. Godage, and D.~G. Caldwell, ``Dynamic continuum arm model for use with underwater robotic manipulators inspired by octopus vulgaris,'' in {\em 2012 IEEE international conference on robotics and automation}, pp.~5289--5294, IEEE, 2012.

\bibitem{stella2023piecewise}
F.~Stella, Q.~Guan, C.~Della~Santina, and J.~Hughes, ``Piecewise affine curvature model: a reduced-order model for soft robot-environment interaction beyond pcc,'' in {\em 2023 IEEE International Conference on Soft Robotics (RoboSoft)}, pp.~1--7, IEEE, 2023.

\bibitem{simo1985finite}
J.~C. Simo, ``A finite strain beam formulation. the three-dimensional dynamic problem. part i,'' {\em Computer methods in applied mechanics and engineering}, vol.~49, no.~1, pp.~55--70, 1985.

\bibitem{rucker_statics_2011}
D.~C. Rucker and R.~J. Webster~III, ``Statics and {Dynamics} of {Continuum} {Robots} {With} {General} {Tendon} {Routing} and {External} {Loading},'' {\em IEEE Transactions on Robotics}, vol.~27, pp.~1033--1044, Dec. 2011.

\bibitem{renda_dynamic_2014}
F.~Renda, M.~Giorelli, M.~Calisti, M.~Cianchetti, and C.~Laschi, ``Dynamic {Model} of a {Multibending} {Soft} {Robot} {Arm} {Driven} by {Cables},'' {\em IEEE Transactions on Robotics}, vol.~30, pp.~1109--1122, Oct. 2014.

\bibitem{gazzola_forward_2018}
M.~Gazzola, L.~H. Dudte, A.~G. McCormick, and L.~Mahadevan, ``Forward and inverse problems in the mechanics of soft filaments,'' {\em Royal Society Open Science}, vol.~5, p.~171628, June 2018.

\bibitem{till_real-time_2019}
J.~Till, V.~Aloi, and C.~Rucker, ``Real-time dynamics of soft and continuum robots based on {Cosserat} rod models,'' {\em The International Journal of Robotics Research}, vol.~38, pp.~723--746, May 2019.

\bibitem{renda_discrete_2018}
F.~Renda, F.~Boyer, J.~Dias, and L.~Seneviratne, ``Discrete {Cosserat} {Approach} for {Multisection} {Soft} {Manipulator} {Dynamics},'' {\em IEEE Transactions on Robotics}, vol.~34, pp.~1518--1533, Dec. 2018.

\bibitem{renda_geometric_2020}
F.~Renda, C.~Armanini, V.~Lebastard, F.~Candelier, and F.~Boyer, ``A {Geometric} {Variable}-{Strain} {Approach} for {Static} {Modeling} of {Soft} {Manipulators} {With} {Tendon} and {Fluidic} {Actuation},'' {\em IEEE Robotics and Automation Letters}, vol.~5, pp.~4006--4013, July 2020.

\bibitem{9272318}
F.~Boyer, V.~Lebastard, F.~Candelier, and F.~Renda, ``Dynamics of continuum and soft robots: A strain parameterization based approach,'' {\em IEEE Transactions on Robotics}, vol.~37, no.~3, pp.~847--863, 2021.

\bibitem{mazzolai2012soft}
B.~Mazzolai, L.~Margheri, M.~Cianchetti, P.~Dario, and C.~Laschi, ``Soft-robotic arm inspired by the octopus: Ii. from artificial requirements to innovative technological solutions,'' {\em Bioinspiration \& biomimetics}, vol.~7, no.~2, p.~025005, 2012.

\bibitem{kier_arrangement_2007}
W.~M. Kier and M.~P. Stella, ``The arrangement and function of octopus arm musculature and connective tissue,'' {\em Journal of Morphology}, vol.~268, pp.~831--843, Oct. 2007.

\bibitem{chang_energy_2023}
H.-S. Chang, U.~Halder, C.-H. Shih, N.~Naughton, M.~Gazzola, and P.~G. Mehta, ``Energy {Shaping} {Control} of a {Muscular} {Octopus} {Arm} {Moving} in {Three} {Dimensions},'' {\em Proceedings of the Royal Society A: Mathematical, Physical and Engineering Sciences}, vol.~479, p.~20220593, Feb. 2023.
\newblock arXiv:2209.04089 [physics].

\bibitem{shih2023hierarchical}
C.-H. Shih, N.~Naughton, U.~Halder, H.-S. Chang, S.~H. Kim, R.~Gillette, P.~G. Mehta, and M.~Gazzola, ``Hierarchical control and learning of a foraging cyberoctopus,'' {\em Advanced Intelligent Systems}, vol.~5, no.~9, p.~2300088, 2023.

\bibitem{kumar_geometrically_2011}
A.~Kumar and S.~Mukherjee, ``A {Geometrically} {Exact} {Rod} {Model} {Including} {In}-{Plane} {Cross}-{Sectional} {Deformation},'' {\em Journal of Applied Mechanics}, vol.~78, p.~011010, Jan. 2011.

\bibitem{tunay_spatial_2013}
I.~Tunay, ``Spatial {Continuum} {Models} of {Rods} {Undergoing} {Large} {Deformation} and {Inflation},'' {\em IEEE Transactions on Robotics}, vol.~29, pp.~297--307, Apr. 2013.
\newblock Conference Name: IEEE Transactions on Robotics.

\bibitem{angles_viper_2019}
B.~Angles, D.~Rebain, M.~Macklin, B.~Wyvill, L.~Barthe, J.~Lewis, J.~Von Der~Pahlen, S.~Izadi, J.~Valentin, S.~Bouaziz, and A.~Tagliasacchi, ``{VIPER}: {Volume} {Invariant} {Position}-based {Elastic} {Rods},'' {\em Proceedings of the ACM on Computer Graphics and Interactive Techniques}, vol.~2, pp.~1--26, July 2019.

\bibitem{antman_theory_2005}
S.~S. Antman, ``Theory of {Rods} {Deforming} in {Space},'' in {\em Nonlinear {Problems} of {Elasticity}}, Applied {Mathematical} {Sciences}, pp.~269--344, New York, NY: Springer, 2005.

\bibitem{geradin2001flexible}
M.~G{\'e}radin and A.~Cardona, ``Flexible multibody dynamics: a finite element approach,'' {\em (No Title)}, 2001.

\bibitem{smith_stable_2018}
B.~Smith, F.~D. Goes, and T.~Kim, ``Stable {Neo}-{Hookean} {Flesh} {Simulation},'' {\em ACM Transactions on Graphics}, vol.~37, pp.~12:1--12:15, Mar. 2018.

\bibitem{linn_geometrically_2013}
J.~Linn, H.~Lang, and A.~Tuganov, ``Geometrically exact {Cosserat} rods with {Kelvin}–{Voigt} type viscous damping,'' {\em Mechanical Sciences}, vol.~4, pp.~79--96, Feb. 2013.

\bibitem{ambrosi_active_2012}
D.~Ambrosi and S.~Pezzuto, ``Active {Stress} vs. {Active} {Strain} in {Mechanobiology}: {Constitutive} {Issues},'' {\em Journal of Elasticity}, vol.~107, pp.~199--212, Apr. 2012.

\bibitem{9486942}
C.~Armanini, M.~Farman, M.~Calisti, F.~Giorgio-Serchi, C.~Stefanini, and F.~Renda, ``Flagellate underwater robotics at macroscale: Design, modeling, and characterization,'' {\em IEEE Transactions on Robotics}, vol.~38, no.~2, pp.~731--747, 2022.

\bibitem{renda_geometrically-exact_2022}
F.~Renda, C.~Armanini, A.~Mathew, and F.~Boyer, ``Geometrically-{Exact} {Inverse} {Kinematic} {Control} of {Soft} {Manipulators} {With} {General} {Threadlike} {Actuators}’ {Routing},'' {\em IEEE Robotics and Automation Letters}, vol.~7, pp.~7311--7318, July 2022.

\bibitem{hairer_geometric_2013}
E.~Hairer, C.~Lubich, and G.~Wanner, {\em Geometric {Numerical} {Integration}: {Structure}-{Preserving} {Algorithms} for {Ordinary} {Differential} {Equations}}.
\newblock Springer {Series} in {Computational} {Mathematics}, Springer Berlin Heidelberg, 2013.

\bibitem{boyer2017poincare}
F.~Boyer and F.~Renda, ``Poincare’s equations for cosserat media: Application to shells,'' {\em Journal of Nonlinear Science}, vol.~27, pp.~1--44, 2017.

\bibitem{hanassy2015stereotypical}
S.~Hanassy, A.~Botvinnik, T.~Flash, and B.~Hochner, ``Stereotypical reaching movements of the octopus involve both bend propagation and arm elongation,'' {\em Bioinspiration \& biomimetics}, vol.~10, no.~3, p.~035001, 2015.

\bibitem{gutfreund1996organization}
Y.~Gutfreund, T.~Flash, Y.~Yarom, G.~Fiorito, I.~Segev, and B.~Hochner, ``Organization of octopus arm movements: a model system for studying the control of flexible arms,'' {\em Journal of Neuroscience}, vol.~16, no.~22, pp.~7297--7307, 1996.

\bibitem{gutfreund_patterns_1998}
Y.~Gutfreund, T.~Flash, G.~Fiorito, and B.~Hochner, ``Patterns of {Arm} {Muscle} {Activation} {Involved} in {Octopus} {Reaching} {Movements},'' {\em The Journal of Neuroscience}, vol.~95, pp.~5976--5987, Aug. 1998.

\bibitem{yekutieli2005dynamic}
Y.~Yekutieli, R.~Sagiv-Zohar, R.~Aharonov, Y.~Engel, B.~Hochner, and T.~Flash, ``Dynamic model of the octopus arm. i. biomechanics of the octopus reaching movement,'' {\em Journal of neurophysiology}, vol.~94, no.~2, pp.~1443--1458, 2005.

\bibitem{wang2022control}
T.~Wang, U.~Halder, E.~Gribkova, M.~Gazzola, and P.~G. Mehta, ``Control-oriented modeling of bend propagation in an octopus arm,'' in {\em 2022 American Control Conference (ACC)}, pp.~1359--1366, IEEE, 2022.

\bibitem{hochner_embodied_2023}
B.~Hochner, L.~Zullo, T.~Shomrat, G.~Levy, and N.~Nesher, ``Embodied mechanisms of motor control in the octopus,'' {\em Current Biology}, vol.~33, pp.~R1119--R1125, Oct. 2023.

\bibitem{golub1969calculation}
G.~H. Golub and J.~H. Welsch, ``Calculation of gauss quadrature rules,'' {\em Mathematics of computation}, vol.~23, no.~106, pp.~221--230, 1969.

\end{thebibliography}
\bibliographystyle{ieeetr}

\end{document}